\documentclass[journal,10pt]{IEEEtran}% {ieeetmlcn}

\usepackage{graphicx,color}
\usepackage{textcomp}
\usepackage{xcolor}
\usepackage{hyperref}
\hypersetup{hidelinks=true}
\usepackage{algorithm,algorithmic}
\def\BibTeX{{\rm B\kern-.05em{\sc i\kern-.025em b}\kern-.08em
    T\kern-.1667em\lower.7ex\hbox{E}\kern-.125emX}}
\AtBeginDocument{\definecolor{tmlcncolor}{cmyk}{0.93,0.59,0.15,0.02}\definecolor{NavyBlue}{RGB}{0,86,125}}

\def\OJlogo{\vspace{-4pt}\includegraphics[height=18pt]{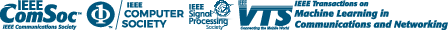}}
\def\seclogo{\vspace{10pt}\includegraphics[height=18pt]{new_logo}}

\def\authorrefmark#1{\ensuremath{^{\textbf{#1}}}}

\usepackage{cite}
\usepackage{amsmath,amssymb,amsfonts}
\usepackage{algorithmic}
\usepackage{lipsum,adjustbox}
\usepackage{verbatim}
\usepackage{graphics}
\usepackage{stackengine}
\usepackage{pgfplots}
\pgfplotsset{width=7cm,compat=1.8}
\usepackage{multirow}
\usepackage{soul}
\usepackage{amssymb}
\usepackage{amsmath}
\usepackage{algorithm}
\usepackage{algorithmic}
\usepackage{graphicx}
\usepackage{subfigure}
\usepackage{textcomp} 
\usepackage{booktabs}
\usepackage{stfloats}
\usepackage{scalefnt}
\usepackage{mathrsfs}       
\DeclareMathAlphabet{\pazocal}{OMS}{zplm}{m}{n}

\usepackage[nolist]{acronym}
    
 \usepackage{bm}

\newcommand{\figref}[1]{Figure~\ref{#1}}

\newcommand{\compl}{\mathbb{C}}         % complex number field, e.g., x \in \compl
\newcommand{\real} {\mathbb{R}}          % real number field, e.g., x \in \real

\newcommand{\ma}  [1]{ \bm{#1} } % matrix (upper case) and vector (lower case)
\newcommand{\set} [1]{{\mathcal {#1}}} % define set
\newcommand{\Kp} {\set{K}_{\text{p}}} % K_on set
\newcommand{\Kd} {\set{K}_{\text{d}}} % K_on set
\newcommand{\Kn} {\set{K}_{\text{n}}} % K_on set
\begin{acronym}
\acro{DSRC}{dedicated short-range communications}
\acro{C-ITS}{cooperative intelligent transport system}
\acro{RSU}{road side unit}
\acro{Bi}{bi-directional}
\acro{TR}{throughput}
\acro{TDL}{tapped delay line}
\acro{SRNN}{Simple RNN}
\acro{ITS}{Intelligent Transportation Systems}
\acro{IEEE}{Institute of Electrical and Electronics Engineers}
\acro{WAVE}{Wireless Access in Vehicular Environment}
\acro{V2V}{vehicle-to-vehicle}
\acro{V2I}{vehicle-to-infrastructure}
\acro{CCH}{control channel}
\acro{SCH}{service channels}
\acro{STS}{short training symbols}
\acro{LTS}{long training symbols}
\acro{SS}{signal symbol}
\acro{SoA}{state-of-the-art}
\acro{DPA}{data-pilot aided}
\acro{STA}{spectral temporal averaging}
\acro{EL}{ensemble learning}
\acro{FL}{frame length}
\acro{CDP}{constructed data pilots}
\acro{TRFI}{time-domain reliable test frequency domain interpolation}
\acro{MMSE-VP}{minimum mean square error using virtual pilots}
\acro{iCDP}{improved CDP}
\acro{SBS}{symbol-by-symbol}
\acro{FBF}{frame-by-frame}
\acro{E-TRFI}{Enhanced TRFI}
\acro{SR-CNN}{super resolution CNN}
\acro{DN-CNN}{denoising CNN}
\acro{RBF}{radial basis function}
\acro{CNN}{convolutional neural network}
\acro{TS-ChannelNet}{temporal spectral ChannelNet}
\acro{WSSUS}{wide-sense stationary uncorrelated scattering}
\acro{TDR}{transmission data rate}
\acro{LSTM}{long short-term memory}
\acro{ALS}{accurate LS}
\acro{SLS}{simple LS}
\acro{ChannelNet}{channel network}
\acro{ADD-TT}{average decision-directed with time truncation}
\acro{WI}{weighted interpolation}
\acro{DD}{decision-directed}
\acro{SR-ConvLSTM}{super resolution convolutional long short-term memory}
\acro{RS}{reliable subcarriers}
\acro{URS}{unreliable subcarriers}
\acro{AE-DNN}{auto-encoder deep neural network}
\acro{AE}{auto-encoder}
\acro{T-DFT}{truncated discrete Fourier transform}
\acro{TA-TDFT}{temporal averaging T-DFT}
\acro{TA}{time averaging}
\acro{PDP}{power delay profile}
\acro{1G}{first generation}
\acro{2G}{second generation}
\acro{3G}{third generation}
\acro{3GPP}{Third Generation Partnership Project}
\acro{4G}{fourth generation}
\acro{5G}{fifth generation}
\acro{802.11}{IEEE 802.11 specifications}
\acro{A/D}{analog-to-digital}
\acro{ADC}{analog-to-digital}
\acro{AM}{amplitude modulation}
\acro{AP}{access point}
\acro{AR}{augmented reality}
\acro{ASIC}{application-specific integrated circuit}
\acro{ASIP}{Application Specific Integrated Processors}
\acro{AWGN}{additive white Gaussian noise}
\acro{BCJR}{Bahl, Cocke, Jelinek and Raviv}
\acro{BER}{bit error rate}
\acro{BFDM}{bi-orthogonal frequency division multiplexing}
\acro{BPSK}{binary phase shift keying}
\acro{BS}{base stations}
\acro{CA}{carrier aggregation}
\acro{CAF}{cyclic autocorrelation function}
\acro{Car-2-x}{car-to-car and car-to-infrastructure communication}
\acro{CAZAC}{constant amplitude zero autocorrelation waveform}
\acro{CB-FMT}{cyclic block filtered multitone}
\acro{CCDF}{complementary cumulative density function}
\acro{CDF}{cumulative density function}
\acro{CDMA}{code-division multiple access}
\acro{CFO}{carrier frequency offset}
\acro{CIR}{channel impulse response}
\acro{CM}{complex multiplication}
\acro{COFDM}{coded-\acs{OFDM}}
\acro{CoMP}{coordinated multi point}
\acro{COQAM}{cyclic OQAM}
\acro{CP}{cyclic prefix}
\acro{CR}{cognitive radio}
\acro{CRC}{cyclic redundancy check}
\acro{CRLB}{Cram\'{e}r-Rao lower bound}
\acro{CS}{cyclic suffix}
\acro{CSI}{channel state information}
\acro{CSMA}{carrier-sense multiple access}
\acro{CWCU}{component-wise conditionally unbiased}
\acro{D/A}{digital-to-analog}
\acro{D2D}{device-to-device}
\acro{DAC}{digital-to-analog}
\acro{DC}{direct current}
\acro{DFE}{decision feedback equalizer}
\acro{DFT}{discrete Fourier transform}
\acro{DL}{deep learning}
\acro{DMT}{discrete multitone}
\acro{DNN}{deep neural network}
\acro{FNN}{feed-forward neural network}
\acro{DSA}{dynamic spectrum access}
\acro{DSL}{digital subscriber line}
\acro{DSP}{digital signal processor}
\acro{DTFT}{discrete-time Fourier transform}
\acro{DVB}{digital video broadcasting}
\acro{DVB-T}{terrestrial digital video broadcasting}
\acro{DWMT}{discrete wavelet multi tone}
\acro{DZT}{discrete Zak transform}
\acro{E2E}{end-to-end}
\acro{eNodeB}{evolved node b base station}
\acro{E-SNR}{effective signal-to-noise ratio}
\acro{EVD}{eigenvalue decomposition}
\acro{FBMC}{filter bank multicarrier}
\acro{FD}{frequency-domain}
\acro{FDD}{frequency-division duplexing}
\acro{FDE}{frequency domain equalization}
\acro{FDM}{frequency division multiplex}
\acro{FDMA}{frequency-division multiple access}
\acro{FEC}{forward error correction}
\acro{FER}{frame error rate}
\acro{FFT}{fast Fourier transform}
\acro{FIR}{finite impulse response}
\acro{FM}		{frequency modulation}
\acro{FMT}{filtered multi tone}
\acro{FO}{frequency offset}
\acro{F-OFDM}{filtered-\acs{OFDM}}
\acro{FPGA}{field programmable gate array}
\acro{FSC}{frequency selective channel}
\acro{FS-OQAM-GFDM}{frequency-shift OQAM-GFDM}
\acro{FT}{Fourier transform}
\acro{FTD}{fractional time delay}
\acro{FTN}{faster-than-Nyquist signaling}
\acro{GFDM}{generalized frequency division multiplexing}
\acro{GFDMA}{generalized frequency division multiple access}
\acro{GMC-CDM}{generalized	multicarrier code-division multiplexing}
\acro{GNSS}{global navigation satellite system}
\acro{GS}{guard symbols}
\acro{GSM}{Groupe Sp\'{e}cial Mobile}
\acro{GUI}{graphical user interface}
\acro{H2H}{human-to-human}
\acro{H2M}{human-to-machine}
\acro{HTC}{human type communication}
\acro{I}{in-phase}
\acro{i.i.d.}{independent and identically distributed}
\acro{IB}{in-band}
\acro{IBI}{inter-block interference}
\acro{IC}{interference cancellation}
\acro{ICI}{inter-carrier interference}
\acro{ICT}{information and communication technologies}
\acro{ICV}{information coefficient vector}
\acro{IDFT}{inverse discrete Fourier transform}
\acro{IDMA}{interleave division multiple access}
\acro{IEEE}{institute of electrical and electronics engineers}
\acro{IF}{intermediate frequency}
\acro{IFFT}{inverse fast Fourier transform}
\acro{IoT}{Internet of Things}
\acro{IOTA}{isotropic orthogonal transform algorithm}
\acro{IP}{internet protocole}
\acro{IP-core}{intellectual property core}
\acro{ISDB-T}{terrestrial integrated services digital broadcasting}
\acro{ISDN}{integrated services digital network}
\acro{ISI}{inter-symbol interference}
\acro{ITU}{International Telecommunication Union}
\acro{IUI}{inter-user interference}
\acro{LAN}{local area netwrok}
\acro{LLR}{log-likelihood ratio}
\acro{LMMSE}{linear minimum mean square error}
\acro{LNA}{low noise amplifier}
\acro{LO}{local oscillator}
\acro{LOS}{line-of-sight}
\acro{LP}{low-pass}
\acro{LPF}{low-pass filter}
\acro{LS}{least squares}
\acro{LTE}{Long Term Evolution}
\acro{LTE-A}{LTE-Advanced}
\acro{LTIV}{linear time invariant}
\acro{LTV}{linear time-variant}
\acro{LUT}{lookup table}
\acro{M2M}{machine-to-machine}
\acro{MA}{multiple access}
\acro{MAC}{multiple access control}
\acro{MAP}{maximum a posteriori}
\acro{MC}{multicarrier}
\acro{MCA}{multicarrier access}
\acro{MCM}{multicarrier modulation}
\acro{MCS}{modulation coding scheme}
\acro{MF}{matched filter}
\acro{MF-SIC}{matched filter with successive interference cancellation}
\acro{MIMO}{multiple-input, multiple-output}
\acro{MISO}{multiple-input single-output}
\acro{ML}{machien learning}
\acro{MLD}{maximum likelihood detection}
\acro{MLE}{maximum likelihood estimator}
\acro{MMSE}{minimum mean squared error}
\acro{MRC}{maximum ratio combining}
\acro{MS}{mobile stations}
\acro{MSE}{mean squared error}
\acro{MSK}{Minimum-shift keying}
\acro{MSSS}[MSSS]	{mean-square signal separation}
\acro{MTC}{machine type communication}
\acro{MU}{multi user}
\acro{MVUE}{minimum variance unbiased estimator}
\acro{NEF}{noise enhancement factor}
\acro{NLOS}{non-line-of-sight}
\acro{NMSE}{normalized mean-squared error}
\acro{NOMA}{non-orthogonal multiple access}
\acro{NPR}{near-perfect reconstruction}
\acro{NRZ}{non-return-to-zero}
\acro{OFDM}{orthogonal frequency division multiplexing}
\acro{OFDMA}{orthogonal frequency division multiple access}
\acro{OOB}{out-of-band}
\acro{OQAM}{offset quadrature amplitude modulation}
\acro{OQPSK}{offset quadrature phase shift keying}
\acro{OTFS}{orthogonal time frequency space}
\acro{PA}{power amplifier}
\acro{PAM}{pulse amplitude modulation}
\acro{PAPR}{peak-to-average power ratio}
\acro{PC-CC}{parallel concatenated convolutional code}
\acro{PCP}{pseudo-circular pre/post-amble}
\acro{PD}{probability of detection}
\acro{pdf}{probability density function}
\acro{PDF}{probability distribution function}

\acro{PFA}{probability of false alarm}
\acro{PHY}{physical layer}
\acro{PIC}{parallel interference cancellation}
\acro{PLC}{power line communication}
\acro{PMF}{probability mass function}
\acro{PN}{pseudo noise}
\acro{ppm}{parts per million}
\acro{PRB}{physical resource block}
\acro{PRB}{physical resource block}
\acro{PSD}{power spectral density}
\acro{Q}{quadrature-phase}
\acro{QAM}{quadrature amplitude modulation}
\acro{QoS}{quality of service}
\acro{QPSK}{quadrature phase shift keying}
\acro{R/W}{read-or-write}
\acro{RAM}{random-access memmory}
\acro{RAN}{radio access network}
\acro{RAT}{radio access technologies}
\acro{RC}{raised cosine}
\acro{RF}{radio frequency}
\acro{rms}{root mean square}
\acro{RRC}{root raised cosine}
\acro{RW}{read-and-write}
\acro{SC}{single-carrier}
\acro{SCA}{single-carrier access}
\acro{SC-FDE}{single-carrier with frequency domain equalization}
\acro{SC-FDM}{single-carrier frequency division multiplexing}
\acro{SC-FDMA}{single-carrier frequency division multiple access}
\acro{SD}{sphere decoding}
\acro{SDD}{space-division duplexing}
\acro{SDMA}{space division multiple access}
\acro{SDR}{software-defined radio}
\acro{SDW}{software-defined waveform}
\acro{SEFDM}{spectrally efficient frequency division multiplexing}
\acro{SE-FDM}{spectrally efficient frequency division multiplexing}
\acro{SER}{symbol error rate}
\acro{SIC}{successive interference cancellation}
\acro{SINR}{signal-to-interference-plus-noise ratio}
\acro{SIR}{signal-to-interference ratio}
\acro{SISO}{single-input, single-output}
\acro{SMS}{Short Message Service}
\acro{SNR}{signal-to-noise ratio}
\acro{STC}{space-time coding}
\acro{STFT}{short-time Fourier transform}
\acro{STO}{symbol time offset}
\acro{SU}{single user}
\acro{AI}{artificial intelligence}
\acro{SVD}{singular value decomposition}
\acro{TD}{time-domain}	
\acro{TDD}{time-division duplexing}
\acro{TDMA}{time-division multiple access}
\acro{TFL}{time-frequency localization}
\acro{TO}{time offset}
\acro{TS-OQAM-GFDM}{time-shifted OQAM-GFDM}
\acro{XAI}{explainable artificial intelligence}
\acro{UE}{user equipment}
\acro{VTV-US}{Vehicle-to-Vehicle urban scenario}
\acro{VTI-US}{Vehicle-to-Infrastructure urban scenario}
\acro{UFMC}{universally filtered multicarrier}
\acro{UL}{uplink}
\acro{US}{uncorrelated scattering}
\acro{USB}{universal serial bus}
\acro{UW}{unique word}
\acro{VLC}{visible light communications}
\acro{VR}{virtual reality}
\acro{WCP}{windowing and \acs{CP}}	\acro{ANN}{artificial neural network}
\acro{GRU}{gated recurrent unit}
\acro{RNN}{recurrent neural network}
\acro{WHT}{Walsh-Hadamard transform}
\acro{WiMAX}{worldwide interoperability for microwave access}
\acro{WLAN}{wireless local area network}
\acro{W-OFDM}{windowed-\acs{OFDM}}	
\acro{WOLA}{windowing and overlapping}	
\acro{WSS}{wide-sense stationary}
\acro{ZCT}{Zadoff-Chu transform}
\acro{ZF}{zero-forcing}
\acro{ZMCSCG}{zero-mean circularly-symmetric complex Gaussian}
\acro{ZP}{zero-padding}
\acro{ZT}{zero-tail}
\end{acronym}

\usepackage[utf8]{inputenc}
\usepackage[english]{babel}
\usepackage{amsthm}
\usepackage{commath}

\newcommand{\Lb}{\pazocal{L}}

\newtheorem{lemma}{Lemma}
\DeclareMathOperator{\dom}{dom}
\usepackage{cleveref}% http://ctan.org/pkg/cleveref
\crefname{lemma}{Lemma}{Lemmas}
\usepackage{flushend}

\begin{document}
% \receiveddate{XX Month, XXXX}
% \reviseddate{XX Month, XXXX}
% \accepteddate{XX Month, XXXX}
% \publisheddate{XX Month, XXXX}
% \currentdate{XX Month, XXXX}
% \doiinfo{TMLCN.2022.1234567}

\markboth{}{Author {et al.}}

\title{Explainable AI for Enhancing Efficiency of DL-based Channel Estimation}

\author{Abdul Karim Gizzini\authorrefmark{1}, Member, IEEE, Yahia Medjahdi\authorrefmark{2}, Member, IEEE, Ali J. Ghandour\authorrefmark{3}\\ and Laurent Clavier\authorrefmark{2}, Senior Member, IEEE\\
\IEEEauthorblockA{\authorrefmark{1} SogetiLabs Research and Innovation (part of Capgemini), F-92130, Issy Les Moulineaux, France\\
%\IEEEauthorrefmark{2} Univ. Lille, CNRS, UMR 8520 - IEMN, F-59000 Lille, France\\
\authorrefmark{2}IMT Nord Europe, Institut Mines T\'el\'ecom, Center for Digital Systems, F-59653 Villeneuve d’Ascq, France\\
\authorrefmark{3}National Center for Remote Sensing - CNRS, Lebanon\\
Email: \ abdul.gizzini@sogeti.fr}
}

% \affil{SogetiLabs Research and Innovation (part of Capgemini), F-92130, Issy Les Moulineaux, France}
% \affil{Centre for Digital Systems, IMT Nord Europe, Institut Mines-Télécom, University of Lille, France}
% \affil{National Center for Remote Sensing, CNRS, Lebanon}
% \affil{Electrical Engineering Department, University of Colorado, Boulder, CO 80309 USA}
% \corresp{Corresponding author: Abdul Karim Gizzini (email: abdul.gizzini@sogeti.com).}

% \authornote{\hl{This paragraph of the first footnote will contain support information, including sponsor and financial support acknowledgment. For example, ``This work was supported in part by the U.S. Department of Commerce under Grant 123456.''}}

\maketitle

\begin{abstract}

The support of artificial intelligence (AI) based decision-making is a key element in future 6G networks. Moreover, AI is widely employed in critical applications such as autonomous driving and medical diagnosis. In such applications, using AI as black-box models is risky and challenging. Hence, it is crucial to understand and trust the decisions taken by these models. Tackling this issue can be achieved by developing explainable AI (XAI) schemes that aim to explain the logic behind the black-box model behavior, and thus, ensure its efficient and safe deployment. Recently, we proposed a novel perturbation-based XAI-CHEST framework that is oriented toward channel estimation in wireless communications. The core idea of the XAI-CHEST framework is to identify the relevant model inputs by inducing high noise on the irrelevant ones. This manuscript provides the detailed theoretical foundations of the XAI-CHEST framework. In particular, we derive the analytical expressions of the XAI-CHEST loss functions and the noise threshold fine-tuning optimization problem. Hence the designed XAI-CHEST delivers a smart input feature selection methodology that can further improve the overall performance while optimizing the architecture of the employed model. Simulation results show that the XAI-CHEST framework provides valid interpretations, where it offers an improved bit error rate performance while reducing the required computational complexity in comparison to the classical DL-based channel estimation.

\end{abstract}



% \maketitle

\section{INTRODUCTION} \label{introduction}

\IEEEPARstart{A}{rtificial} intelligence (AI) is expected to play a crucial role in the overall design of future networks. In particular, 6G will transform the classical Internet of Things (IoT) to ``connected intelligence",
by leveraging the power of AI to connect billions of devices and systems worldwide~\cite {ref_6G}. This concept is defined as native  (AI) which is a key element that differentiates 6G networks from the previous wireless networks. In native AI, distributed AI will be embedded within the functionality of all layers~\cite{ref_nativeAI} to support demands for high data rates and low latency-critical applications. 

Generally speaking, the AI-enabled intelligent architecture for 6G networks defines several layers including the intelligent sensing layer~\cite{ref_6G_2}. It is worth mentioning that robust environment monitoring and data detection are of great interest in 6G smart applications like autonomous driving~\cite{ref_latency}. Note that ensuring the reliability of the intelligent sensing layer is highly impacted by the channel estimation accuracy since a precisely estimated channel response influences the follow-up equalization and decoding operations at the receiver, therefore, it affects the sensing accuracy~\cite{ref_survey}.
In this context, channel estimation is one of the major physical (PHY) layer issues due to the doubly-selective nature of the channel in mobile applications. Conventional channel estimation schemes such as \ac{LS} ignores the presence of noise in the estimation process and requires the transmission of a large number of pilots which decreases the transmission data rate. Whereas, the \ac{LMMSE} channel estimator provides good performance assuming the prior knowledge of the channel and noise statistics in addition to its high computational complexity.

\subsection{DL-BASED CHANNEL ESTIMATION}

Recently, \ac{DL} has been employed in the PHY layer of wireless communications~\cite{ref_ai_phy_1}, including channel estimation~\cite{ref_AE_DNN, ref_STA_DNN, ref_TRFI_DNN, ref_wi_cnn}, due to its ability in providing good performance-complexity trade-offs. Among different {\ac{DL}} networks, {\acp{FNN}} have been widely used as a post-processing unit following conventional channel estimators. In~\cite{ref_DL_Chest3}, the authors proposed an end-to-end FNN-based scheme for channel estimation and signal detection, where it directly detects the received bits from the received signal. The proposed FNN model consists of $3$ hidden layers with $500$, $250$, and $120$ neurons, respectively. We note that in this scheme the \ac{FNN} model is trained to predict 16 bits only, hence, several concatenated models are needed according to the total number of transmitted bits. Using the same \ac{FNN} model proposed in~\cite{ref_DL_Chest3}, the authors in~\cite{ref_DL_Chest2} proposed an \ac{FNN}-based channel estimation scheme that is used to predict the channel response using the received signal, received pilots, and previously estimated channel. Simulation results show that using the previously estimated channel as an \ac{FNN} input improves the channel estimation accuracy. Another \ac{FNN}-based channel estimation scheme has been proposed in~\cite{ref_DL_Chest1}, where \ac{LS} channel estimation is first applied and combined with the previously estimated channel to be fed as an input to a $3$ hidden layer \ac{FNN} consisting of $512$, $256$, and $128$ neurons, respectively. As reported in~\cite{ref_DL_Chest1}, employing \ac{LS} as an \ac{FNN} input improves the channel estimation accuracy and provides a comparable performance to the \ac{LMMSE} channel estimation scheme. 

To further improve the performance while preserving low computational complexity, the authors in~\cite{ref_AE_DNN, ref_STA_DNN, ref_TRFI_DNN} tried a different strategy that is based on improving the conventional channel estimation accuracy and employing low complex \ac{FNN} models as post-processing units. In~\cite{ref_AE_DNN}, the authors proposed an \ac{FNN}-based channel estimation scheme that applies {\ac{DPA}} channel estimation on top of {\ac{LS}} channel estimation. After that a $3$ hidden layer \ac{FNN} consisting of $40$, $20$, and $40$ neurons, respectively is utilized. Simulation results reveal that improving the initial channel estimation allows the use of a low-complex \ac{FNN} model while recording a significant performance improvement in comparison to the conventional channel estimation schemes. Similarly in~\cite{ref_STA_DNN} and~\cite{ref_TRFI_DNN} the authors proposed two \ac{FNN}-based channel estimation schemes that employ {\ac{STA}}~\cite{ref_STA} and {\ac{TRFI}}~\cite{ref_TRFI} channel estimation before the \ac{FNN} model which consist of $3$ hidden layers with $15$ neurons each. \ac{STA}-\ac{FNN} and \ac{TRFI}-\ac{FNN}  outperform the \ac{DPA}-\ac{FNN}~\cite{ref_AE_DNN} while recording a substantial computational complexity decrease.

In addition to {\ac{FNN}} models, {\ac{RNN}} and {\ac{CNN}} models have been also used within the channel estimation by also trying to combine several inputs such as the received signal, pilots, and initially estimated channel. {\ac{RNN}}-based channel estimation schemes~\cite{ref_RNN_paper, ref_DL_CHEST_LSTM,ref_BiRNN} can provide better channel tracking capability in comparison to the \ac{FNN}-based channel estimation. Whereas, {\ac{CNN}}-based channel estimation~\cite{ref_CNN5,ref_CNN6} is used in the frame-by-frame channel estimation, where the previous, current, and future pilots are employed in the channel estimation for
each received signal. Thus, improving the channel estimation accuracy with the cost of a higher computational complexity as well as inducing high processing time in comparison to the \ac{RNN} and \ac{FNN}-based channel estimation. We note that in this work we focus on the FNN-based channel estimation since we are targeting low-complex low-latency DL-based solutions.

\begin{figure*}
    \centering
\includegraphics[width=2\columnwidth]{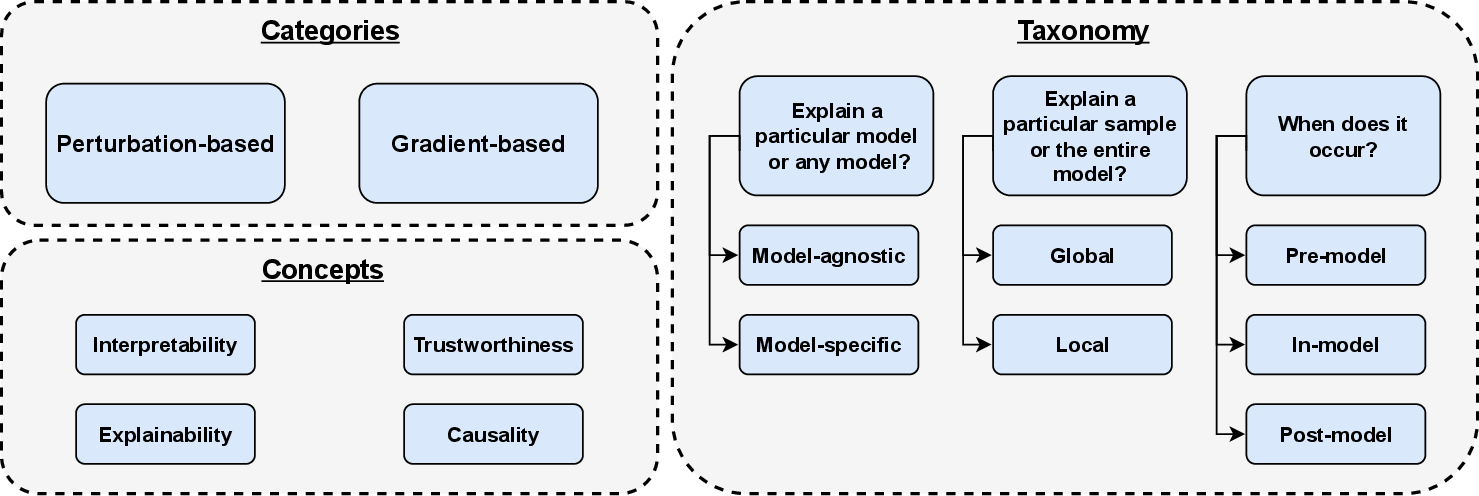}
    \caption{XAI categories, concepts, and taxonomy.}
    \label{fig:xai_con_tax}
\end{figure*}

There exist three main issues concerning the discussed channel estimation schemes which can be defined as follows: (\textit{i}) Identifying the relevant inputs: as previously discussed, the majority of \ac{DL}-based channel estimation schemes use a combination of information as an input to the utilized \ac{DL} model without any clear criteria. Hence, \textbf{is there a way to better select %optimize 
the DL model inputs?} (\textit{ii}) High computational complexity: to guarantee good performance, highly complex \ac{DL} architectures are employed. However, motivated by the fact that low-complex architectures are required in low-latency applications, so \textbf{is there a real need for such high-complex architectures?} (\textit{iii}) Trustworthiness: Despite the good generalization and performance abilities offered by different {\ac{DL}}-based channel estimation schemes, they lack trustworthiness since they are considered as ``black box" models. Consequently, researchers and industrial leaders are not able to trust the employment of these models in real-case sensitive applications~\cite{ref_trust_XAI}. Therefore, \textbf{is there a way to provide interpretability to the decision-making strategy employed {\ac{DL}} black box models?}

The mentioned issues can be tackled by developing {\ac{XAI}} schemes that provide a reasonable explanation of the decisions taken by black-box models. Thus, ensuring the transparency of the employed models by transforming them from black-box into white-box models that can be safely employed in practice. In the following paragraphs, we will discuss the main \ac{XAI} concepts, taxonomy, and deployment within the wireless communication research domain.

\subsection{\MakeUppercase{XAI Main Concepts, Categories, and Taxonomy}}

\ac{XAI} defines four main concepts as shown in~\figref{fig:xai_con_tax}: (\textit{i}) Interpretability: is based on the model design and it refers to how much the black-box model can be understood by humans. For example, decision tree models have good interpretability since a human can easily understand their logic. \textit{ii}) Explainability: is the ability to clarify and justify a particular prediction performed by the black-box model. Hence, it aims to clarify the internal functioning of the employed model. (\textit{iii}) Trustworthiness: is the ability to make professionals feel confident in the decisions taken by the black-box model. (\textit{iv}) Causality: is related to the generalization ability of the black-box model, where models should be able to detect cause-effect relations and adapt to environmental changes. 

Generally speaking, \ac{XAI} methods can be divided into two main categories~\cite{nielsen2022robust}: (\textit{i}) Perturbation-based or gradient-free methods, where the concept is to perturb input features by masking or altering their values and record the effect of these changes on the model performance. (\textit{ii}) Gradient-based methods where the gradients of the output are calculated with respect to the input via back-propagation and used to estimate importance scores of the input features. Moreover, in terms of the provided explanations, the \ac{XAI} methods can be further classified into~\cite{arya2019one}: 

\begin{itemize}
    \item Model-agnostic vs model-specific: Model-agnostic
\ac{XAI} schemes are independent of the internal architecture of the black-box model including the weights and the hidden layers. Whereas, model-specific schemes depend on a specific model like {\ac{FNN}} or {\ac{CNN}} and can not be generalized to any other model. Therefore, model-agnostic schemes are characterized by their high flexibility and can be used despite the type of the considered model.

    \item Local vs global: Local \ac{XAI} schemes are those that generate explanations for a group of samples, thus, they are highly dependent on the utilized dataset. In contrast, global \ac{XAI} schemes generate explanations that are related more to the model behavior.  

    \item Pre-model, in-model, and post-model strategies: Knowing that the \ac{XAI} schemes can be applied throughout the entire development pipeline. Hence, interpretability can be acquired in three main phases. Pre-modeling explainability is used to define the useful features of the dataset for a better representation. Hence, pre-modeling aims to perform exploratory data analysis, explainable feature engineering, and dataset description.  In contrast, in-model explainability is to develop inherently explainable models instead of generating black box models. Finally, the post-model explainability method extracts explanations that are dependent on the model predictions.

\end{itemize}

\vspace{-0.3cm}

\subsection{\MakeUppercase{XAI for Wireless Communications}}

Wireless communications are still in the early stages of using \ac{XAI}. The majority of related works in the literature are surveys and reviews about the guidelines and importance of using \ac{XAI} in wireless communications. In~\cite{ref_xai_6G_1} the authors provided a review of the core concepts of \ac{XAI} including definitions and possible performance vs. explainability trade-offs. They mainly focused on reviewing the recent \ac{DL}-based schemes for the PHY and MAC layers and specified the explainability level of the studied schemes which is in general low. In~\cite{ref_xai_6G_2} the authors proposed a novel \ac{XAI} knowledge-powered framework for network automation
that effectively adapts to the dynamic changes of complex communication systems as well as provides a human-understandable explanation. The proposed \ac{XAI} scheme aims to explain the decision-making for automated path selection within the network.

The deployment of \ac{XAI} in the open radio access network (O-RAN) was recently surveyed in~\cite{ref_xai_6G_ran_1}, where the authors performed a comprehensive survey on the use of \ac{XAI} to design a trustworthy and explainable O-RAN architecture. Moreover, an explainable machine learning operations (MLOps) for streamlined automation-native 6G networks has been proposed in~\cite{ref_xai_6G_ran_2,ref_xai_6G_ran_3}, where Shapley additive explanations (SHAP) XAI scheme is employed to assign the features importance~\cite{lundberg2017unified}. We note that SHAP XAI scheme has been also employed for short-term
resource reservation in 5G networks~\cite{ref_xai_res_alloc_1} and energy-efficient resource allocation, where the problem becomes more challenging~\cite{ref_xai_res_alloc_2, ref_xai_res_alloc_3}. It is worth mentioning that, the majority of \ac{DL}-based resource allocation schemes are based on deep
reinforcement learning (DRL) where SHAP assigns importance to the features used by the DRL agent at each state. These features could be the number of active antennas, utilized bandwidth, number of connected users, and the average data rate. We note that SHAP is useful in such applications, however, it can not be efficiently used in channel estimation due to the absence of predefined features and the high dimensionality of the \ac{DL} input vector.

In addition to network optimization and resource allocation, \ac{XAI} has been employed also in internet-of-things (IoT) networks. In~\cite{ref_xai_iot_1} the authors presented a comprehensive survey on \ac{XAI}
solutions for IoT systems including the state-of-the-art past and ongoing
research activities. In particular, they focused on the \ac{XAI} for IoT adaptive solutions using several architectures based on 5G services, cloud services, and big data management. In~\cite{ref_xai_3} the authors proposed a novel model-agnostic \ac{XAI} scheme denoted as 
transparency relying upon statistical theory (TRUST) for numerical
applications. They further tested the proposed TRUST scheme on cybersecurity of the industrial IoT (IIoT). Simulation results show that
TRUST scheme outperforms the local interpretable model-agnostic explanations (LIME) scheme~\cite{ribeiro2016should} in terms of performance, speed, and explainability.

\vspace{-0.3cm}

\subsection{\MakeUppercase{Motivation and Contributions}}

To the best of our knowledge, the methodology of deploying XAI schemes in PHY layer applications, specifically, channel estimation is still unclear. Noting that the proposed XAI-based schemes for network optimization~\cite{ref_xai_6G_ran_2}, resource allocation~\cite{ref_xai_res_alloc_1}, and secured IoT~\cite{ref_xai_iot_1} can not be adapted to the PHY layer applications because in such applications there are no clear discriminative features within the model inputs. In this context, this paper aims to design a novel {\ac{XAI}} framework for FNN-based channel estimation, denoted as XAI-CHEST. This framework is based on a perturbation-based model-agnostic global pre-model methodology that jointly performs the channel estimation task and provides the corresponding interpretability. The XAI-CHEST concept has been partially proposed in~\cite{10368353}, where the key idea is to provide the interpretability of black box models by inducing noise on the model input while preserving accuracy. The model inputs are then classified into relevant and irrelevant sets based on the induced noise. The contributions of this work can be summarized as follows:

\begin{itemize}
 
    \item Establishing the theoretical foundations of the XAI-CHEST framework where the detailed loss functions are formulated. 

    \item Deriving the analytical expression and the corresponding simulations of the noise threshold optimization to select the best threshold used in filtering the relevant model inputs.

    \item Showing that using only relevant inputs instead of the full set improves the performance of the considered \ac{DL}-based channel estimators.

    \item Optimizing the architecture of the considered DL-based channel estimator where minimizing the relevant model inputs resulted in a reduction of the model's hidden layers while preserving performance levels. 

\end{itemize}

The remainder of this paper is organized as follows: Section~\ref{system_model} presents the system model in addition to the  {\ac{DL}}-based channel estimators to be interpreted. Section~\ref{proposed_xai_scheme} illustrates the designed XAI-CHEST framework as well as the noise threshold fine-tuning optimization problem. In Section~\ref{simulation_results}, the performance of the designed XAI-CHEST framework in terms of \ac{BER} is analyzed considering several criteria. Finally, Section~\ref{conclusions} concludes the manuscript.

\textbf{Notations}: Throughout the paper, vectors are defined
with lowercase bold symbols $\ma{s}$. The ($i,k$) element of ${\ma{s}}$ is represented by ${\ma{s}}_{i}[k]$, where $i$ and $k$ stand for the OFDM symbol and the subcarrier indices, respectively. We note that the full OFDM symbol ${\ma{s}}_{i} \in \compl^{K \times 1}$ includes ${\ma{s}}_{i,d} \in \compl^{K_{d} \times 1}$ data symbols and ${\ma{s}}_{i,p} \in \compl^{K_{p} \times 1}$ pilots.

\section{\MakeUppercase{System Model}} \label{system_model}
 
This section illustrates the considered generic system model in addition to the considered DL-based channel estimation scheme to be interpreted.

\subsection{\MakeUppercase{OFDM transceiver}}

In this work, we consider single-input and single-output (SISO) {\ac{OFDM}}-based transmission with non-linear radio frequency (RF) represented by the high power amplifier (HPA) at the {\ac{OFDM}} transmitter. As shown in~{\figref{fig:TxRx}}, the first operation on the transmitter side is the binary bits generation. Generated bits are scrambled in order to randomize the bits pattern, which may contain
long streams of $1$s or $0$s. The scrambled bits are then passed to the encoder, which introduces some redundancy into the bits stream. This redundancy is used for error correction that allows the receiver to combat the effects of the channel, hence reliable communications can be achieved.

Bits interleaving is used to cope with the channel noise such as burst errors or fading. The interleaver rearranges input bits such that consecutive bits are split among different
blocks. This can be done using a permutation process that ensures that adjacent bits are modulated onto non-adjacent subcarriers and thus allows better error correction at the receiver. After that, the interleaved bits are mapped according to the employed modulation technique, i.e., BPSK, QPSK, 16QAM, 64QAM, etc. Bits mapping operation is followed by constructing the \ac{OFDM} symbols to be transmitted. The data symbols and pilots are mapped to the active subcarriers and passed to the IFFT block to generate the time-domain \ac{OFDM} symbols and followed by inserting the \ac{CP}. Finally, the CP-OFDM symbol is subjected to the impacts of HPA non-linear distortion as well as the channel and the {\ac{AWGN}} noise.

At the receiver side, the \ac{CP} is removed and the FFT applied to the received symbol. Channel estimation and equalization are performed where the equalized data are de-mapped to obtain the encoded bits. Afterwards, deinterleaving, decoding, and descrambling are performed to obtain the detected bits. We note that the employed system model is based on the IEEE 802.11p standard~{\cite{ref_IEEE_Spec}}.

\subsection{\MakeUppercase{Signal Model}}

Consider a frame consisting of $I$ {\ac{OFDM}} symbols.  The $i$-th transmitted frequency-domain {\ac{OFDM}} symbol ${\ma{s}}_i \in \compl^{K\times1}$, can be expressed as:
\begin{equation}
   {\ma{s}}_i[k] = \left\{
            \begin{array}{ll}        
                {\ma{s}}_{{i,d}}[k],&\quad k \in \Kd \\
                {\ma{s}}_{{i,p}}[k],&\quad k \in \Kp \\
                0,&\quad k \in \Kn \\
            \end{array}\right.
\label{eq: xK}
\end{equation}
where $0 \leq k \leq K - 1$ denotes the subcarrier index. We note that $K_{\text{on}}$ useful subcarriers are used where $K_{\text{on}} = K_{p} + K_{d}$.  $ {\ma{s}}_{{i,p}} \in \compl^{K_{p} \times 1}$ and $ {\ma{s}}_{{i,d}} \in \compl^{K_{d} \times 1}$ represent the allocated pilot symbols and the 
modulated data symbols at a set of subcarriers denoted $\Kp$ and $\Kd$, respectively. $K_{n}$
$=K-K_{\text{on}}$ denotes the null guard band subcarriers. $K_{\text{cp}}$ samples are added to the time-domain OFDM symbol resulting in $ {\ma{x}}_{{i}} \in \compl^{K + K_{\text{cp}} \times 1}$ which is then passed to the HPA. According to the Bussgang theorem~\cite{bussgang1952crosscorrelation}, the  HPA output $ {\ma{u}}^{\prime}_{{i}} \in \compl^{K + K_{\text{cp}} \times 1}$ can be expressed as follows:

\begin{equation}
	\begin{split}
		{\ma{u}}^{\prime}_{{i}}
		&= \rho {\ma{x}}_i + {\ma{z}}^{\prime}_{i},
	\end{split}            
	\label{eq: sm1}
\end{equation}

where $\rho$ and ${\ma{z}}^{\prime}_{i}$ refer to the complex gain and the non-linear distortion (NLD), respectively. After that $\rho$ is compensated at the transmitter and ${\ma{u}}^{\prime}_{{i}}$ can be rewritten as:

\begin{equation}
	\begin{split}
		{\ma{u}}_{{i}} = \frac{{\ma{u}}^{\prime}_{{i}}}{\rho}
		&= {\ma{x}}_i + {{\ma{z}}}_{i},
	\end{split}            
	\label{eq: sm2}
\end{equation}

where ${{\ma{z}}_i} = \frac{{\ma{z}}^{\prime}_i}{\rho}$ denotes the remaining NLD of the HPA. The received frequency-domain {\ac{OFDM}} symbol ${\ma{y}}_{{i}} \in \compl^{K_{\text{on}}\times1}$ is expressed as follows:

\begin{figure}[t]
\includegraphics[width=0.5\textwidth]{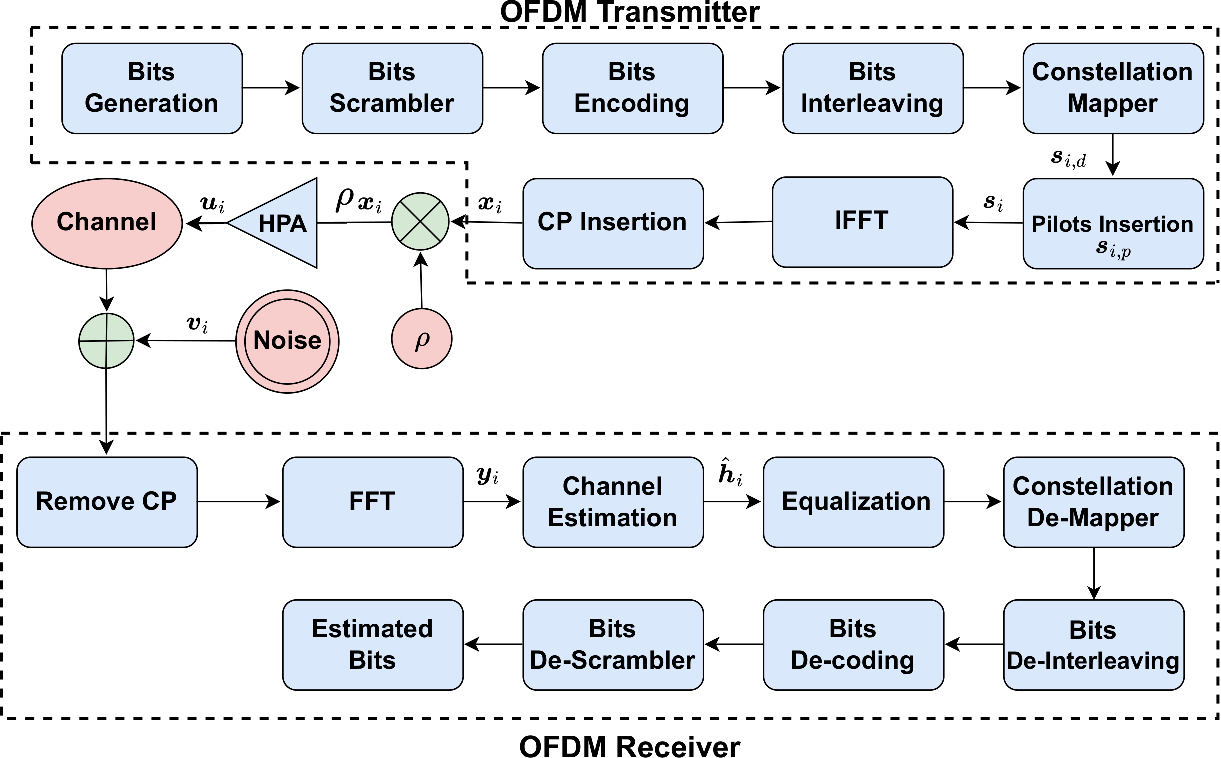}
	\caption{OFDM transmitter-receiver block diagram.}
	\label{fig:TxRx}
\end{figure}

\begin{equation}
	\begin{split}
		{\ma{y}}_{{i}}[k] 
		&= {\ma{h}}_i[k] \tilde{\ma{u}}_i[k] + {\ma{e}}_{i}[k] + \tilde{\ma{v}}_i[k],
  %,~ k \in \Kon + \tilde{\ma{e}}_{i}[k] 
	\end{split}            
	\label{eq: system_model}
\end{equation}
where 
\begin{equation}
     \tilde{\ma{u}}_i[k] = \left\{
            \begin{array}{ll}        
                {\ma{s}}_{{i}}[k],&\quad \text{linear RF} \\
                {\ma{s}}_{{i}}[k] + \tilde{\ma{z}}_{{i}}[k] ,&\quad \text{non-linear RF} \\
            \end{array}\right.
\label{eq: sm4}
\end{equation}

${\ma{h}}_i \in \compl^{K_{\text{on}} \times 1}$, ${\tilde{\ma{v}}}_i  \in \compl^{K_{\text{on}} \times 1}$, and ${\ma{e}}_{i}  \in \compl^{K_{\text{on}} \times 1}$ refer to the frequency-time response of the doubly-selective channel, the {\ac{AWGN}} at the $i$-th {\ac{OFDM}}, and the Doppler-induced inter-carrier interference, respectively. ${\ma{e}}_{i}$ can be expressed as:

\begin{equation}
	\begin{split}
		{\ma{e}}_{i}[k] &=  \frac{1}{{{K}}} \sum_{\substack{q=0 \\ q \neq k}}^{{K}-1} \sum_{n=0}^{{K}-1} {\ma{h}}_i[q,n] e^{-j2\pi\frac{ n (k-q)}{{K}}} {\ma{s}}_i[q].
		\label{eq: Doppler}
	\end{split}
\end{equation}

The full derivation of  ${\ma{e}}_{i}$ can be found in~\cite{9743925}. We note that the objective is to obtain the channel frequency response estimate denoted by $\hat{{\ma{h}}}_i \in \compl^{K_{\text{on}} \times 1}$.

\subsection{\MakeUppercase{DL-based Channel Estimation}}

Conventional channel estimation depends highly on environment conditions. In frequency-selective slow fading channels, the preamble-based channel estimation is sufficient, since the communication system encounters only muti-path fading and the channel is not changing over time. However, in double selective channels, the impact of Doppler interference is added to the communication system. Thus, the estimated channel at the beginning of the frame, i.e., the preambles, becomes outdated and channel tracking becomes more challenging, especially, in high mobility scenarios. To cope with this challenge, pilot subcarriers are allocated within a transmitted {\ac{OFDM}} symbol to allow better channel tracking over time, where several conventional channel estimation and tracking schemes are proposed in the literature. In order to further improve the conventional channel estimation accuracy, \ac{DL} models are applied as post-processing on top of conventional channel estimators. In this work, we considered the {\ac{STA}}-{\ac{FNN}} channel estimator~\cite{ref_STA_DNN} as a case study, where we used the optimized XAI-CHEST framework to provide the corresponding reasonable interpretations.   

Conventional {\ac{STA}} channel estimation scheme  ~\cite{ref_STA} is based on the {\ac{DPA}} estimation where the demapped data subcarriers of the previously received {\ac{OFDM}} symbol are used to estimate the channel for the current {\ac{OFDM}} symbol such that:

\begin{equation}
{\ma{d}}_i =  \mathfrak{D} \big( \frac{{\ma{y}}_i}{\hat{{\ma{h}}}_{\text{DPA}_{i-1}}}\big)
,~ \hat{{\ma{h}}}_{\text{DPA}_{0}} = \hat{{\ma{h}}}_{\text{LS}},
\label{eq: DPA_1}
\end{equation}
where $\mathfrak{D}(.)$ refers to the demapping operation to the nearest constellation point according to the employed modulation order. $\hat{{\ma{h}}}_{\text{LS}}$ stands for the LS estimated channel at the received preambles. Thereafter, the {\ac{DPA}} channel estimates are updated in the following manner: 
\begin{equation}
\hat{{\ma{h}}}_{\text{DPA}_{i}} = \frac{{\ma{y}}_i}{{\ma{d}}_i}.
\label{eq: DPA_2}
\end{equation}

After that, frequency-domain averaging is applied where the {\ac{DPA}} estimated channel at each subcarrier is updated as follows:

\begin{equation}
\hat{{\ma{h}}}_{\text{FD}_{i}}[k] = \sum_{\lambda = -\beta}^{\lambda = \beta} \omega_{\lambda} \hat{{\ma{h}}}_{\text{DPA}_{i}}[k + \lambda], ~ \omega_{\lambda} = \frac{1}{2\beta+1}.
\label{eq: STA_4}
\end{equation}

Finally, time-domain averaging is employed to reduce the AWGN noise impact such that:

\begin{equation}
\hat{{\ma{h}}}_{\text{STA}_{i}} = (1 - \frac{1}{\alpha})  \hat{{\ma{h}}}_{\text{STA}_{i-1}} + \frac{1}{\alpha}\hat{{\ma{h}}}_{\text{FD}_{i}}.
\label{eq: STA_5}
\end{equation}

We note that conventional {\ac{STA}} channel estimation performs well in the low {\ac{SNR}} region. However, it suffers from a considerable error floor in high {\ac{SNR}} regions due to the large {\ac{DPA}} demapping error resulting from~\eqref{eq: DPA_1} and the fixed frequency and time averaging coefficients $\alpha = \beta = 2$ in~\eqref{eq: STA_4} and~\eqref{eq: STA_5}, respectively. Therefore, the conventional {\ac{STA}} channel estimation scheme is not practical in real-case scenarios due to the high doubly-selective channel variations. As a workaround, a $3$ hidden layer {\ac{FNN}} consisting of $15$ neurons per layer is utilized as a nonlinear post-processing unit following {\ac{STA}}. As shown in~\cite{ref_STA_DNN}, STA-FNN can better capture the frequency correlations of the channel samples, in addition to correcting the conventional {\ac{STA}} estimation error. 

\section{\MakeUppercase{XAI-CHEST Framework Design}} \label{proposed_xai_scheme}

\begin{figure*}[t]
\centering
\includegraphics[width=2\columnwidth]{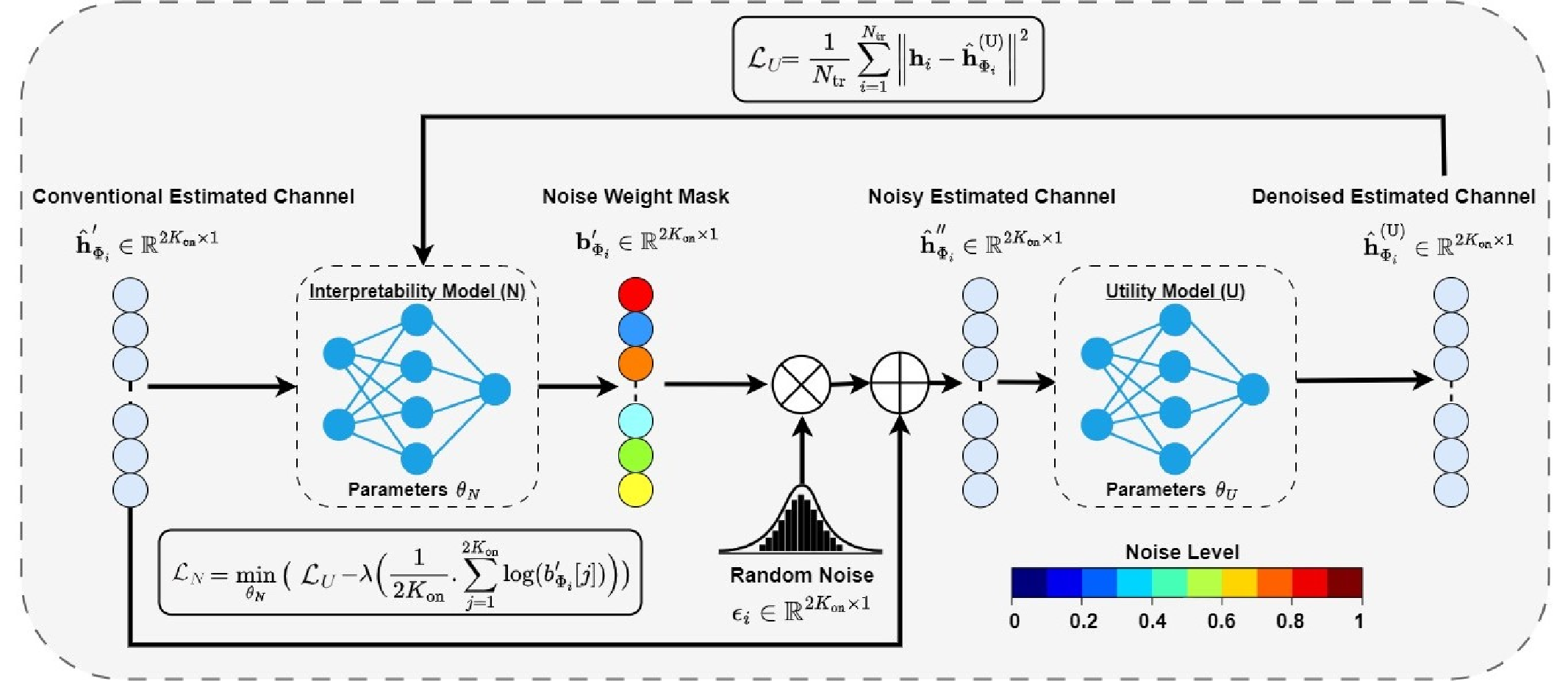}
\caption{Block diagram of the XAI-CHEST framework. The first step is to train the $U$ model and freeze its parameters $\theta_U$. After that, the N model is trained where the objective is to boost the growth of $\ma{b}_{\Phi_{i}}^{\prime}$ while preserving the same performance of the pre-trained $U$ model~{\eqref{eq:lossn}}. Finally, the subcarriers are filtered based on $\ma{b}_{\Phi_{i}}^{\prime}$, where higher noise weight signifies that the corresponding subcarrier is irrelevant (red colors). In contrast, low noise weights mean that the subcarriers are relevant (blue colors). We recall that in this work we are considering the STA-FNN channel estimation scheme, hence, $\Phi = \text{STA}$.}\label{fig:proposed_xai_scheme}
\end{figure*}

Providing external interpretability of the black-box model used for channel estimation could be achieved through classifying the model's input into relevant and irrelevant by employing a perturbation-based methodology. The main intuition is that if a subcarrier is relevant for the decision-making of a trained black box model, then adding noise with high weight to this subcarrier would negatively impact the accuracy of the model. Whereas, if the subcarrier is not contributing to the decision-making of the model, then whatever the induced noise is, the channel estimation accuracy will be preserved. Therefore, it is expected that considering only the relevant subcarriers as model inputs would improve channel estimation performance in comparison to the case where the full subcarriers are given to the model. Moreover, by reducing the model input size, the employed architecture could be further optimized resulting in significantly decreasing the overall computational complexity. Hence, by using the XAI-CHEST framework, we can obtain a reasonable interpretation of the model decision-making methodology, improve the channel estimation performance as well as reduce the required computational complexity.

\subsection{\MakeUppercase{Methodology}}

Let $U$ be the black-box utility model with parameters $\theta_U$. In general, the $U$ model refers to the channel estimation model, where $\hat{{\ma{h}}}^{\prime}_{\Phi_{i}}  \in \real^{2K_{\text{on}} \times 1}$ and $\hat{{\ma{h}}}^{(\text{U})}_{\Phi_{i}} \in \real^{2K_{\text{on}} \times 1}$ denote the conventional estimated channel that is given as an input to the $U$ model and the corresponding output, respectively.  We note that the size of $\hat{{\ma{h}}}^{\prime}_{\Phi_{i}}$ is $2K_{\text{on}}$ since the conventional estimated channel is converted from complex to real domain before further processing by stacking the real and imaginary values vertically in one vector. $\Phi$ refers to the employed conventional channel estimation scheme. The objective is to provide a reasonable interpretation of the behavior of the $U$ model. Besides the $U$ model, we define the interpretability noise model $N$, with parameters $\theta_N$, whose purpose is to compute the weight of the noise induced to each subcarrier within the $U$ input vector. The key idea is that the induced noise weights of the $N$ model should not impact the accuracy of the $U$ model. This could be achieved by customizing the loss function of the $N$ model that will adjust the induced noise while simultaneously maximizing the performance of the $U$ model. We note that the $U$ model is trained before the \ac{XAI} processing of the $N$ model, i.e., the weights of the $U$ model are frozen. Moreover, the $U$ and $N$ models have the same {\ac{FNN}} architecture.

\begin{figure}[t]
    \centering
\includegraphics[width=1\columnwidth]{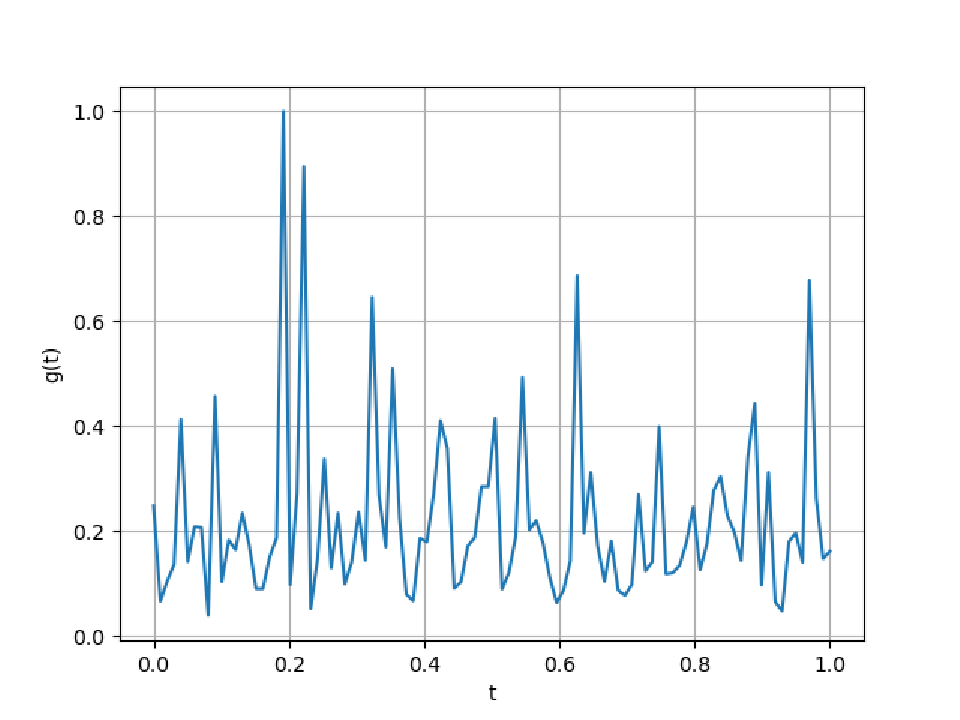}
    \caption{Normalized restricted loss function.}
    \label{fig:optimization}
\end{figure}

Let $\hat{{\ma{h}}}^{\prime}_{\Phi_{i}}$ be the input of the interpretability $N$ model. The role of the $N$ model is to find a mask $\ma{b}^{\prime}_{\Phi_{i}} \in \mathbb{R}^{2 K_{\text{on}} \times 1}$ that can be represented as follows:

\begin{equation}
   \ma{b}^{\prime}_{\Phi_{i}} = N(\hat{{\ma{h}}}^{\prime}_{\Phi_{i}}, \theta_{N}),
\end{equation}
where $\ma{b}^{\prime}_{\Phi_{i}} = (b^{\prime}_{\Phi_{i}}[1], b^{\prime}_{\Phi_{i}}[2], ..., b^{\prime}_{\Phi_{i}} [2 K_{\text{on}}]) \in [0,1]^{{{2K_{\text{on}}}}}$ determines the weight (standard deviation) of the noise applied to each element in $\hat{{\ma{h}}}^{\prime}_{\Phi_{i}}$. We note that the scaling of $ \ma{b}^{\prime}_{\Phi_{i}}$ is achieved using the sigmoid activation function. After that, the generated noise weight mask $\ma{b}^{\prime}_{\Phi_{i}}$ is first multiplied by a random noise $\epsilon \sim  \mathcal{N} (0,1)$ sampled from the standard normal distribution, the resultant is added to the conventional estimated channel vector, such that:

\begin{equation}
  \hat{{\ma{h}}}^{\prime\prime}_{\Phi_{i}} = \hat{{\ma{h}}}^{\prime}_{\Phi_{i}} + \ma{b}_{\Phi_{i}}^{\prime} \epsilon.
\end{equation}

After that, $\hat{{\ma{h}}}^{\prime\prime}_{\Phi_{i}}$ is fed as input to the $U$ model, such that:

\begin{equation}
   \hat{{\ma{h}}}^{(\text{U})}_{\Phi_{i}} = U(\hat{{\ma{h}}}^{\prime\prime}_{\Phi_{i}}, \theta_{U}).
\end{equation}

The customized loss function of the $N$ model can be expressed as follows:

\begin{equation}
\Lb_{N} = \min_{\theta_{N}}  \big( \Lb_{U} - \lambda \Lb_{X}\big),
\label{eq:lossn}    
\end{equation}

$\Lb_{U}$ denotes the loss unction of the $U$ model when $\hat{{\ma{h}}}^{\prime\prime}_{\Phi_{i}}$ is fed as an input. Hence, $\Lb_{U}$ can be expressed as: 

\begin{equation}
\Lb_{U} = \frac{1}{N_{tr}}. \sum_{i = 1}^{N_{tr}} \left\| {{\ma{h}}}_{{i}} - \hat{{\ma{h}}}^{(\text{U})}_{\Phi_{i}} \right\| ^{2},
\label{eq:lossu}
\end{equation}
where ${{\ma{h}}}_{{i}}$ refers to the true channel and $N_{tr}$ is the number of training samples. Moreover, the induced noise is controlled by $\Lb_{X}$ that can be written as:

\begin{equation}
\Lb_{X} = \frac{1}{2K_{\text{on}}}. \sum_{j = 1}^{2K_{\text{on}}} \log(b^{\prime}_{\Phi_{i}}[j])
\label{eq:obj2}
\end{equation}

We would like to mention that the objective of $\Lb_{N}$ is to keep the loss of the $U$ model as low as possible. In other words, minimizing the added noise by the $N$ model while maximizing the generated $\ma{b}^{\prime}_{\Phi_{i}}$. It is worth mentioning that the interpretability measure aims to find a maximum number of low-significant elements. Hence, the term $-\Lb_{X}$ gives a negative value, which will get closer to zero when more weights are close to one, meaning that our $N$ model finds more irrelevant elements and can better highlight the significant features. We note that $\lambda$ is a parameter that allows to give more or less weight to the interpretability measure. The problem is similar to a minimization on $\Lb_{U}$ with a constraint on $\Lb_{X}$ ($\lambda$ being a Lagrange multiplier) or simply a regularization term.

Finally, in the testing phase, $\ma{b}_{\Phi_{i}}^{\prime}$ is scaled back to $\ma{b}_{\Phi_{i}} \in \mathbb{R}^{K_{\text{on}} \times 1}$, where the noise weight of the real and imaginary parts for each subcarrier are averaged. The motivation behind this averaging lies in the fact that it is noticed during the training phase that the interpretability model produces almost the same noise weight for the real and imaginary parts of each subcarrier within $\hat{{\ma{h}}}^{\prime}_{\Phi_{i}}$. The block diagram of the XAI-CHEST framework and the $N$ model training procedure are illustrated in~\figref{fig:proposed_xai_scheme} and algorithm~{\ref{algo:noise_training}}, respectively. We note that $\hat{{\ma{H}}}^{\prime}_{\Phi} \in \mathbb{R}^{2 K_{\text{on}} \times I_{\text{tr}}} $ and ${{\ma{H}}} \in \mathbb{R}^{2 K_{\text{on}} \times I_{\text{tr}}} $ denote the training dataset pairs of the conventional estimated channels and the true ones, where $I_{\text{tr}}$ is the size of the training dataset.

\begin{algorithm}
\caption{$N$ model training}
\begin{algorithmic} 
\REQUIRE 
 Conventional estimated channel: $\hat{{\ma{H}}}^{\prime}_{\Phi}$, true channel: ${{\ma{H}}}$, learning rate: $\eta$, trained $U$ model with
parameters $\theta_U$
\ENSURE Trained N model with parameters $\theta_N$
\WHILE{not converged}
\FOR{$\hat{{\ma{h}}}^{\prime}_{\Phi_{i}} \in \hat{{\ma{H}}}^{\prime}_{\Phi_{i}}$, ${{\ma{h}}}_{{i}} \in {{\ma{H}}}_{{i}}$}    
\STATE { $\ma{b}_{\Phi_{i}}^{\prime}$ $\gets$ $N(\hat{{\ma{h}}}^{\prime}_{\Phi_{i}}, \theta_{N})$}
\STATE {$\epsilon$ $\gets$ $\mathcal{N} (0,1)$}
\STATE {$\hat{{\ma{h}}}^{\prime\prime}_{\Phi_{i}}$ $\gets$ $\hat{{\ma{h}}}^{\prime}_{\Phi_{i}} + \ma{b}_{\Phi_{i}}^{\prime} \epsilon$}
\STATE {$\hat{{\ma{h}}}^{(\text{U})}_{\Phi_{i}}$ $\gets$ $ U(\hat{{\ma{h}}}^{\prime\prime}_{\Phi_{i}}, \theta_{U})$}
\STATE{ $\Lb_{U}$ $\gets$ $ \text{MSE} \big( {{\ma{h}}}_{{i}} - \hat{{\ma{h}}}^{(\text{U})}_{\Phi_{i}} 
\big)$}
\STATE{ $\Lb_{X}$ $\gets$ $ \text{mean} \big( \log(\ma{b}^{\prime}_{\Phi_{i}}) \big)$}
\STATE{ $\Lb_{N}$ $\gets$ $\Lb_{U}$ $ - \lambda \Lb_{X}$}
\STATE {$\theta_N$ $\gets$ $\theta_N +  \eta \frac{\partial \Lb_{N} }{\partial \theta_N }$ } 
\ENDFOR
\ENDWHILE
\end{algorithmic}
\label{algo:noise_training}
\end{algorithm}

\begin{table*}[t]
\renewcommand{\arraystretch}{1.6}
\centering
\caption{Characteristics of the employed channel models following Jake's Doppler spectrum.}
\label{tb:VCMC}
\begin{tabular}{|c|c|c|}
\hline
\textbf{Channel model} & \textbf{Average path gains {[}dB{]}}                                                                                       & \textbf{Path delays {[}ns{]}}                                                                            \\ \hline
VTV-EX                 & \begin{tabular}[c]{@{}c@{}}{[}0, 0, 0, -6.3, -6.3, -25.1, -25.1, -25.1, -22.7, -2.27, -22.7{]}\end{tabular}     & \begin{tabular}[c]{@{}c@{}}{[}0, 1, 2, 100, 101, 200, 201, 202, 300, 301, 302{]}\end{tabular} \\ \hline
VTV-SDWW             & \begin{tabular}[c]{@{}c@{}}{[}0, 0, -11.2, -11.2, -19, -21.9, -25.3, -25.3, -24.4,-28, -26.1, -26.1{]}\end{tabular} & \begin{tabular}[c]{@{}c@{}}{[}0, 1, 100, 101, 200, 300, 400, 401, 500, 600, 700, 701{]}\end{tabular} \\ \hline
\end{tabular}
\end{table*}

\subsection{\MakeUppercase{Noise Weight Threshold Fine Tuning}}

After accomplishing the $N$ model training, the fine-tuning of the noise weight threshold denoted by $\gamma$ is essential for classifying the model inputs into relevant and irrelevant. This could be formulated as an optimization problem, where the objective is to select the best input combination that minimizes the \ac{MSE} between the corresponding estimated channel by the $U$ model and the true channel. We note that the fine-tuning optimization problem is subjected to improving or preserving the {\ac{BER}} in comparison to the case where the full subcarriers are given to the $U$ model. 

Let $\Omega$ be the generic input given to the $U$ model and $\Psi$ be the optimized model input. The considered fine-tuning optimization problem can be mathematically expressed as:

\begin{equation}
\begin{aligned}
\min_{\Psi, \theta_{U}} \quad &  \Lb_{U} = \frac{1}{N_{tr}}. \sum_{i = 1}^{N_{tr}} \big( {\tilde{\ma{h}}}_{{i}} - U( \Omega = \Psi, \theta_{U}) \big)^{2}\\
\textrm{s.t.} \quad & \text{BER} (\Omega = \Psi) \leq \text{BER} (\Omega = \hat{\tilde{\ma{h}}}^{\prime\prime}_{\text{STA}_{i}})  \\
\end{aligned}
\label{eq:threshold_fine_tuning}
\end{equation}

We note that the defined optimization problem in~{\eqref{eq:threshold_fine_tuning}} is not convex. The non-convexity can be shown by the line restriction method illustrated in~{\cref{lemma1}}~\cite{boyd2004convex}.

\begin{lemma}\label{lemma1}
\textbf{Restriction of a convex function to a line} \\
A function \(f\): \(\mathbb{R}^{n}\rightarrow \mathbb{R}\) is convex, if and only if \\ \(\forall x \in \dom f\) and \(\forall v\in \mathbb{R}^{n}\), the function \(g = f(x + tv)\) is convex on \(\dom g = \{ t\in \mathbb{R}\; |\; x + tv \in \dom f \} \) 
\end{lemma}

{\cref{lemma1}} is based on the line restriction method to prove the convexity of the considered function. In this context, the initial loss function $\Lb_{U}$ is reduced to the restricted loss function denoted as $g(t) = \Lb_{U} (\theta_{U} + tv) $, where  $v$ and $t$ denote the randomly selected slice and the step size, respectively. Figure~{\ref{fig:optimization}} shows the $g(t)$ where we can see numerous local minima signifying the non-convexity of our optimization function expressed in~{\eqref{eq:threshold_fine_tuning}}. We note that in the next section, we provided a heuristic solution of~{\eqref{eq:threshold_fine_tuning}}, where the BER vs noise weight threshold is analyzed and the best threshold is selected according to the lowest recorded BER among all the considered thresholds. 

\section{\MakeUppercase{Simulation Results}} \label{simulation_results}

\begin{figure*}[t]
    \centering
    \subfigure[Noise weight distribution]{\includegraphics[width=0.32\textwidth]{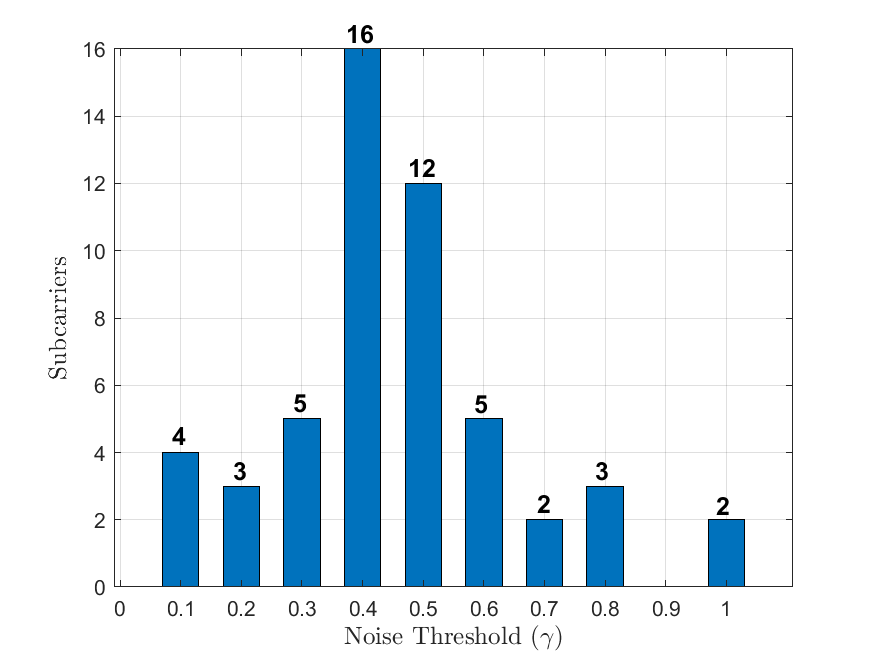}} 
    \subfigure[LFS channel model - QPSK modulation ]{\includegraphics[width=0.32\textwidth]{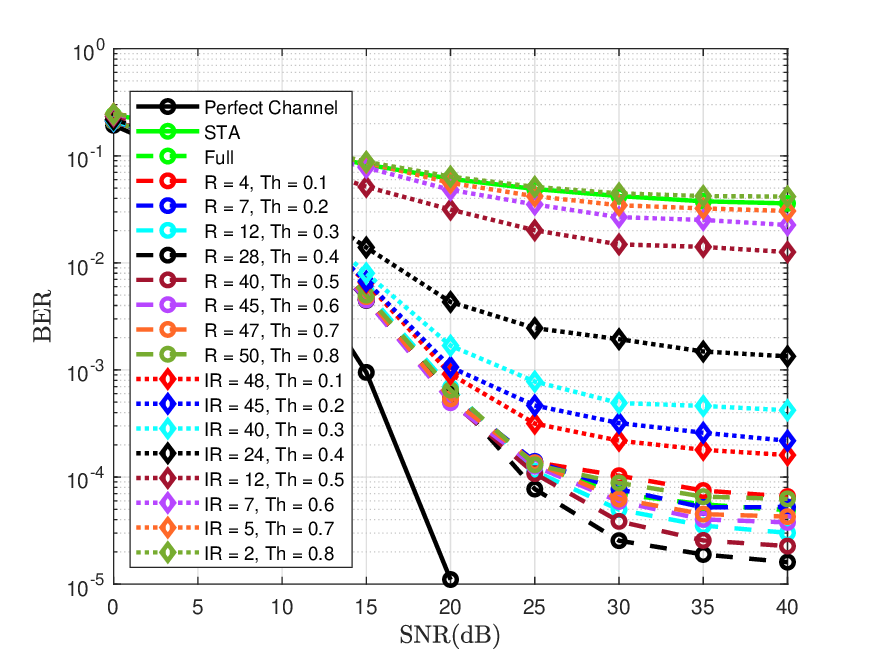}} 
    \subfigure[BER performance vs. noise threshold]{\includegraphics[width=0.32\textwidth]{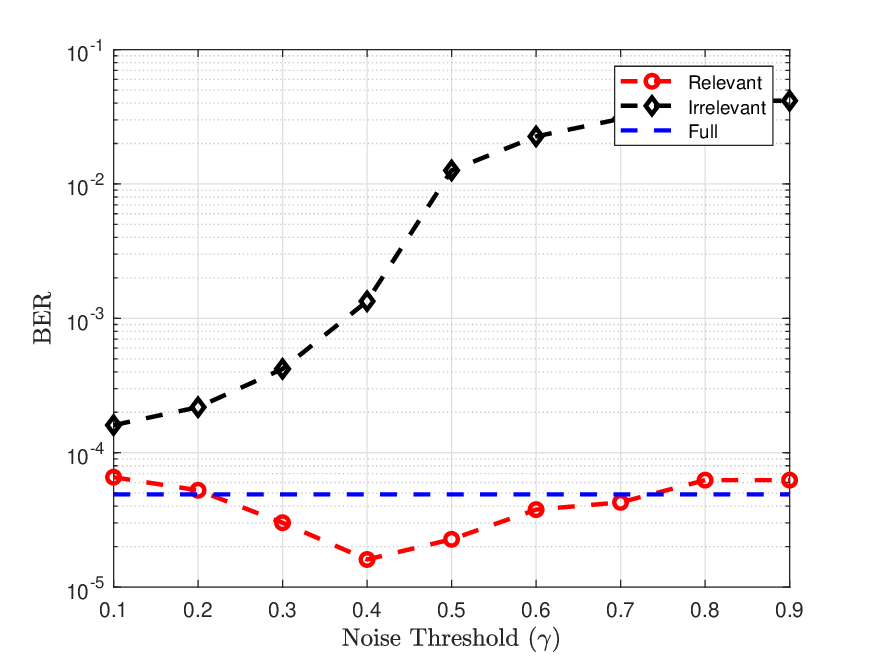}}
    \caption{Fine-tuning of the noise threshold $\gamma$ considering the HFS channel model and QPSK modulation.}
    \label{fig:A}
\end{figure*}

This section illustrates the performance evaluation of the proposed XAI-CHEST framework, where {\ac{BER}} performance of STA-FNN channel estimation scheme is analyzed taking into consideration full, relevant, and irrelevant subcarriers. 
First of all, we start with the noise weight threshold fine-tuning, where the simulation-based solution of~\eqref{eq:threshold_fine_tuning} is provided. After that, the performance evaluation is performed according to several criteria including the (\textit{i}) modulation order, (\textit{ii}) frequency selectivity of the channel, (\textit{iii}) training \ac{SNR}, and (\textit{iv}) conventional channel estimation accuracy. Finally, a detailed computational complexity analysis is discussed where we show that further significant reduction in the overall computational complexity can be achieved by employing only the relevant subcarriers identified by the proposed XAI-CHEST framework. We note that the considered channel models~\cite{r19} are shown in Table~{\ref{tb:VCMC}}: (\textit{i}) Low-frequency selectivity (LFS), where VTV Expressway (VTV-EX) scenario is employed. (\textit{ii}) High-frequency selectivity (HFS), where VTV Expressway Same Direction with Wall (VTV-SDWW) scenario is considered. In both scenarios, Doppler frequency $f_d = 1000$ Hz is considered. 
Both the $U$ and $N$ models are trained using a $100,000$ OFDM symbols dataset, splitted into $80\%$ training, and $20\%$ testing. ADAM optimizer is used with a learning rate $lr = 0.001$ with batch size equals $128$ for $500$ epoch. Simulation parameters are based on the IEEE 802.11p standard~{\cite{ref_survey}}, where the comb pilot allocation is used so that $K_{p} = 4$, $K_{d} = 48$, $K_{n} = 12$, and $I = 50$. Table~\ref{tb:DNN_LSTM_params} summarizes the simulation parameters considered in this work. Finally, we note that the STA channel estimation is considered as an initial estimation prior to the FNN processing. Hence, $\Phi = \text{STA}$, unless stated otherwise.

\begin{table}
 \renewcommand{\arraystretch}{1.5}
	\setlength{\tabcolsep}{3pt}
	\centering
	\caption{Parameters of the studied STA-FNN channel estimation scheme.}
	\label{tb:DNN_LSTM_params}
	\begin{tabular}{|c|c|}
	\hline
		\textbf{Parameter}      & \textbf{Values}                    \\
				\hline
		STA-\ac{FNN} (Hidden layers; Neurons per layer) & (3;15-15-15)  \\ \hline
		Activation function              & ReLU                     \\ \hline
		Number of epochs        & 500                                \\ \hline
		Training samples        & 800000                             \\ \hline
		Testing samples        & 200000                             \\ \hline
		Batch size          & 128                                    \\ \hline
		Optimizer       & ADAM                                       \\ \hline
		Loss function      & MSE                                     \\ \hline
		Learning rate        & 0.001                                 \\ \hline
	    Training SNR        & 40 dB                                 \\ \hline
	\end{tabular}
\end{table}

\subsection{\MakeUppercase{Noise weight threshold fine-tuning}\label{noise_thr}}

The first step of solving~\eqref{eq:threshold_fine_tuning} is to analyze the distribution of the noise weight vector $\ma{b}^{\prime}_{\text{STA}_{i}}$ over the input subcarriers. Both the $U$ and $N$ models are trained using the HFS channel model with QPSK modulation and $40$ dB training SNR. We note that we train the models on  {\ac{SNR}} = $40$ dB since the models can learn the channel better when the training is performed at a high SNR value because the impact of the channel is higher than the impact of the AWGN noise in this SNR range~\cite{ref_LS_DNN}. Owing to the robust generalization properties of DL, trained networks can still estimate the channel even if the AWGN noise increases, i.e., at low SNR values. 
%Therefore, the training of the models is performed using {\ac{SNR}} = $40$ dB to attain the best performance.

Figure~{\ref{fig:A}}(a) shows the distribution of $\ma{b}^{\prime}_{\text{STA}_{i}}$. We notice that the majority of subscribers are distributed more towards zero. This signifies that the model is not sure if the subcarriers can be neglected or not. It is worth mentioning that the pilot subcarriers are assigned the lowest noise weight, i.e., $0.1$ which reveals that the $U$ model is not able to neglect the estimated channels at the pilots, and considering them is crucial for high estimation accuracy. This is consistent with the channel estimation rules, where the channel estimates at the pilots are very close to the ideal channel estimation.

Selecting the optimized $U$ model input $\Psi$ is determined by choosing the best noise weight threshold $\gamma$ which is mainly responsible for classifying the subcarriers into relevant and irrelevant. To select the optimal $\gamma$, we simulated the {\ac{BER}} considering all possible values, i.e, $\gamma = [0.1, 0.2, 0.3, ..., 0.8]$. In each case, we trained the $U$ model considering both the relevant and irrelevant subcarrier combinations.

As shown in Figure~{\ref{fig:A}}(b), we can notice that considering $\gamma = 0.4$ gives the best {\ac{BER}} performance among other thresholds. Therefore, the STA-FNN model needs only $|\Psi| = 28$ subcarriers out of the full set, i.e,  $|\Psi| = 52$ in order to provide the best possible performance in the considered HFS channel model. On the contrary, all the irrelevant subcarrier combinations are worse than the full case in terms of {\ac{BER}} performance. In other words, considering $|\Psi| = 48$ which corresponds to excluding only the four pilot subcarriers is not enough to preserve the BER performance of the full case.

\begin{figure*}[t]
    \centering
    \subfigure[HFS channel model - QPSK modulation]{\includegraphics[width=0.32\textwidth]{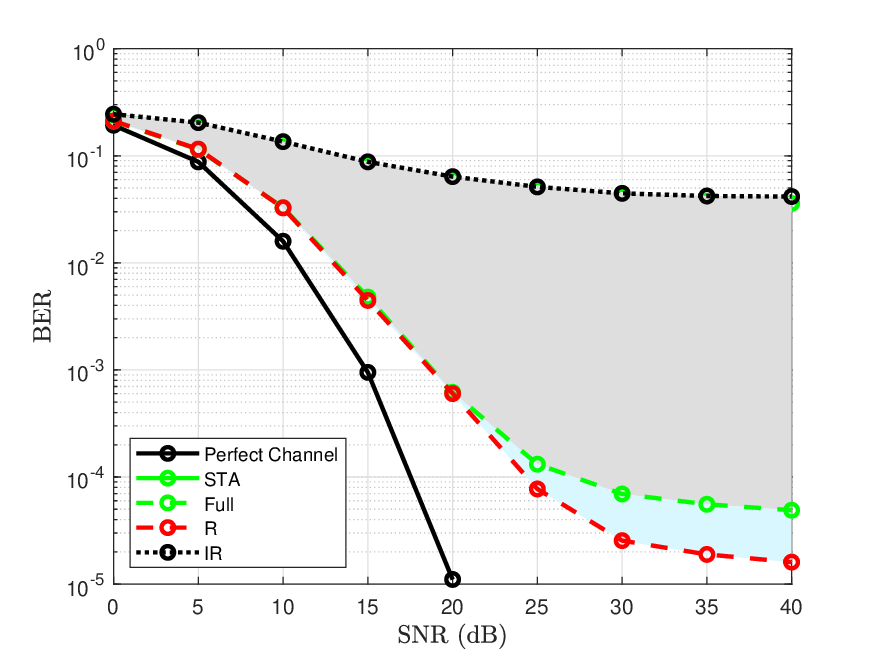}} 
    \subfigure[HFS channel model - 16QAM modulation]{\includegraphics[width=0.32\textwidth]{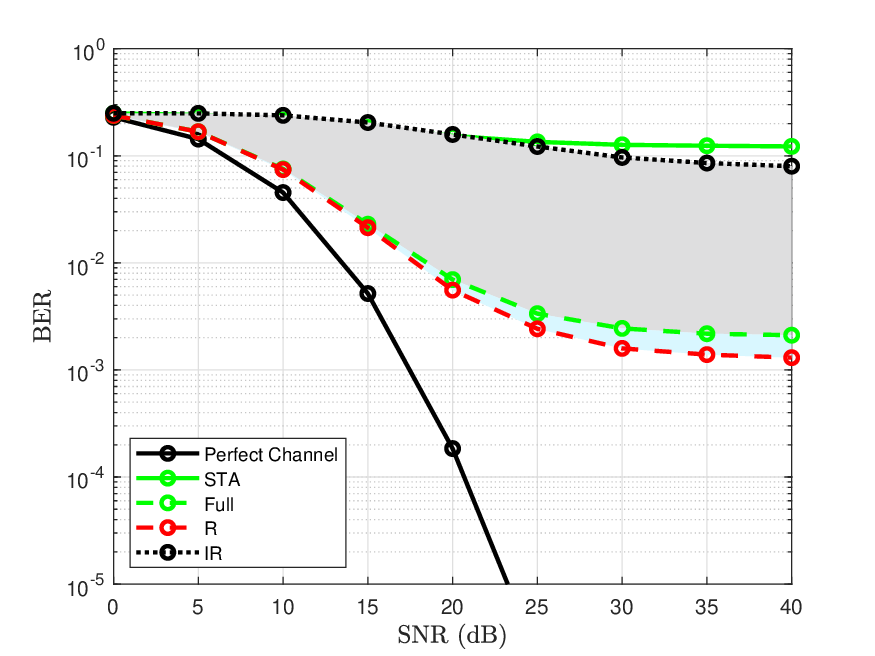}} 
    \subfigure[HFS channel model - 64QAM modulation]{\includegraphics[width=0.32\textwidth]{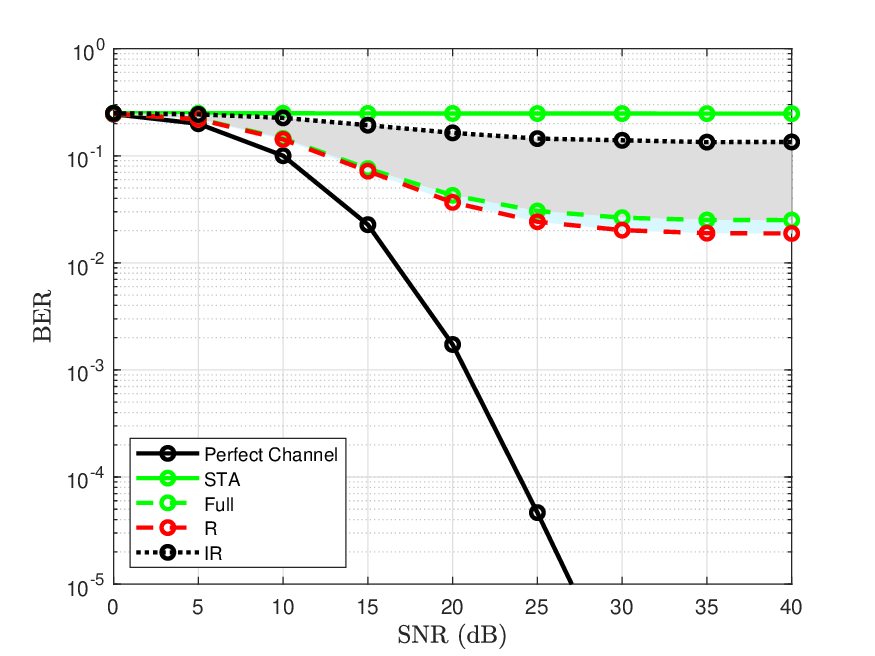}}
    \caption{BER considering HFS channel model employing QPSK, 16QAM, and 64QAM, respectively. We note that the highlighted gray and blue areas refer to the BER performance when considering the irrelevant and relevant subcarriers, respectively.To provide better reading, the best and worst relevant and irrelevant BER performances are only shown, where the regions between the irrelevant-full and full-relevant are highlighted in gray and blue, respectively.}
    \label{fig:C2}
\end{figure*}

\begin{figure*}[t]
    \centering
    \subfigure[Noise weight distribution]{\includegraphics[width=0.49\textwidth]{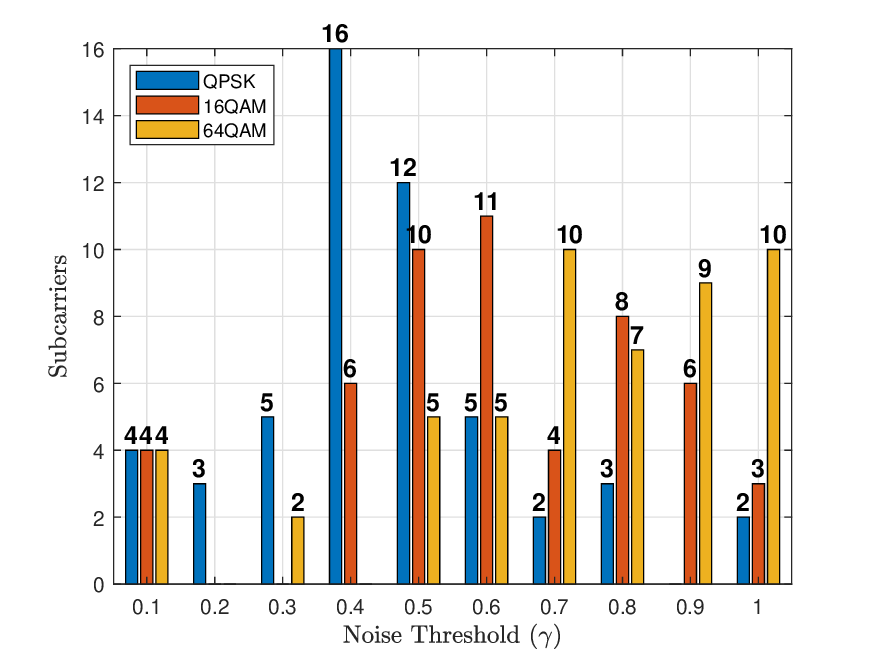}} 
    \subfigure[BER performance vs. noise threshold]{\includegraphics[width=0.49\textwidth]{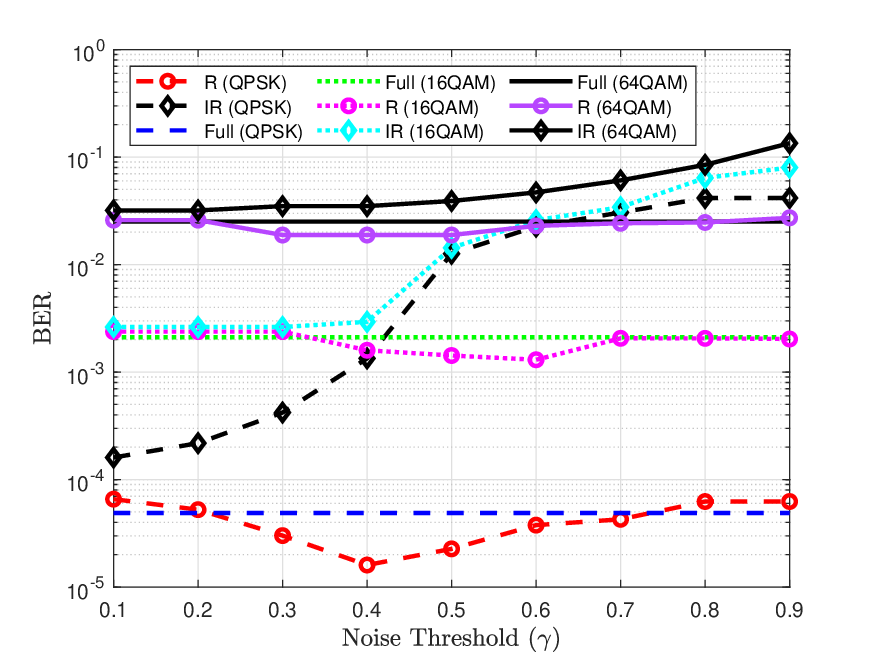}} 
    \caption{Noise distribution and BER threshold analysis for HFS channel model employing QPSK, 16QAM, and 64QAM, respectively. We note that R, IR, and Full in (b) refer to employing only the selected relevant, irrelevant, and Full subcarriers as model inputs, respectively.}
    \label{fig:C1}
\end{figure*}

Figure~{\ref{fig:A}}(c) shows the {\ac{BER}} in terms of $\gamma$ considering SNR = $40$ dB. Again, increasing $\Psi$ is beneficial until reaching  $\gamma = 0.4$, where the BER performance degrades. This signifies that in complicated scenarios as the case in employing the HFS channel model, the proposed perturbation-based XAI scheme can smartly filter out the relevant model inputs which maximize the its performance.

\subsection{\MakeUppercase{Impact of modulation order} \label{mod_order}}

In this section, we further investigate the impact of the employed modulation order on the noise weight distribution considering also the HFS channel model. Figure~\ref{fig:C2} shows the BER performance of employing the HFS channel model using QPSK, 16QAM, and 64QAM, respectively. In general, the BER performance degrades as the modulation order increases. This degradation is mainly due to the impact of the dominant multi-path fading in addition to the DPA remapping error. Moreover, in this scenario, employing only the four pilot subcarriers performs almost similarly to the full case. To improve further the BER performance, more relevant subcarriers are needed. Therefore, when the environment becomes more challenging, the channel variation increases among the subcarriers, thus, the noise distribution is shifted towards zero signifying the need for more relevant subcarriers.

\begin{figure*}[t]
    \centering
    \subfigure[LFS channel model - QPSK modulation]{\includegraphics[width=0.32\textwidth]{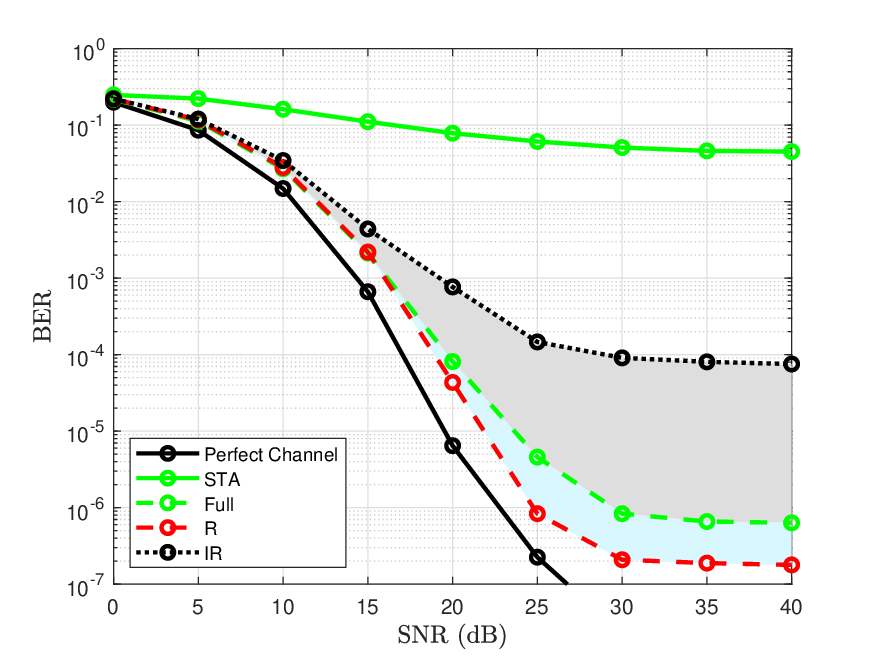}} 
    \subfigure[LFS channel model - 16QAM modulation]{\includegraphics[width=0.32\textwidth]{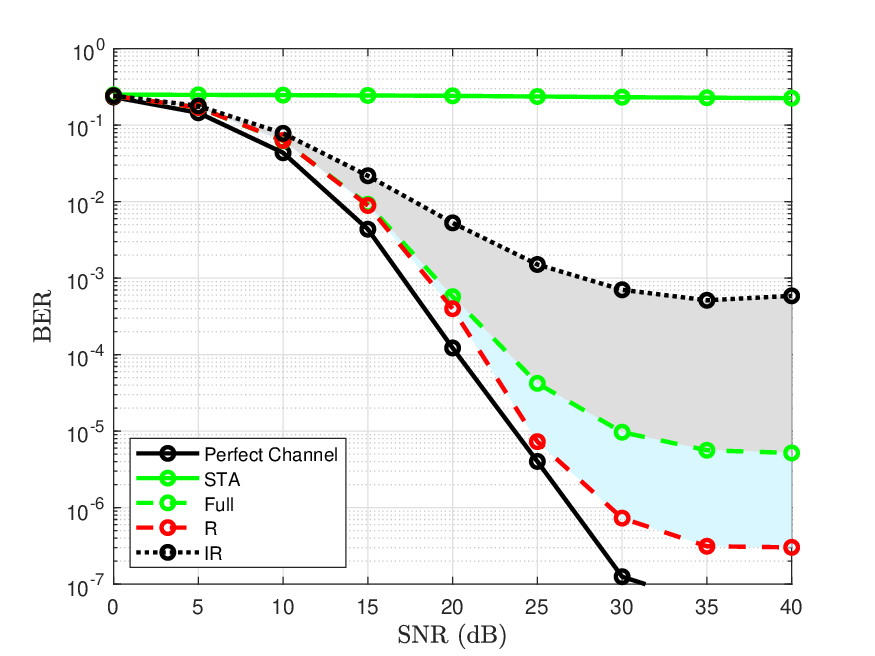}} 
    \subfigure[LFS channel model - 64QAM modulation]{\includegraphics[width=0.32\textwidth]{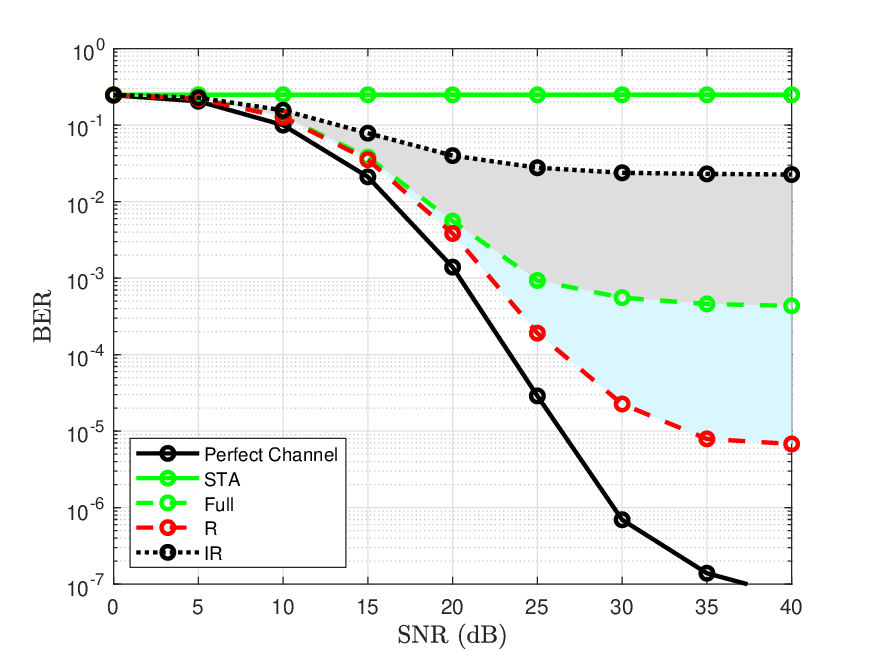}}
    \caption{BER considering LFS channel model employing QPSK, 16QAM, and 64QAM, respectively. We note that the highlighted gray and blue areas refer to the BER performance when considering the irrelevant and relevant subcarriers, respectively. }
    \label{fig:B2}
\end{figure*}

\begin{figure*}[t]
    \centering
    \subfigure[Noise weight distribution]{\includegraphics[width=0.49\textwidth]{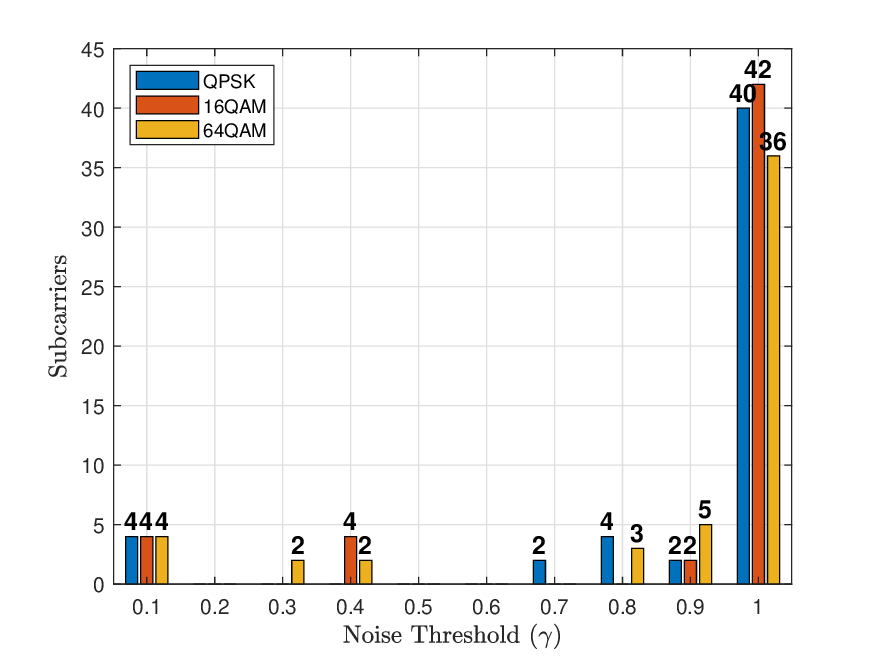}} 
    \subfigure[BER performance vs. noise threshold]{\includegraphics[width=0.49\textwidth]{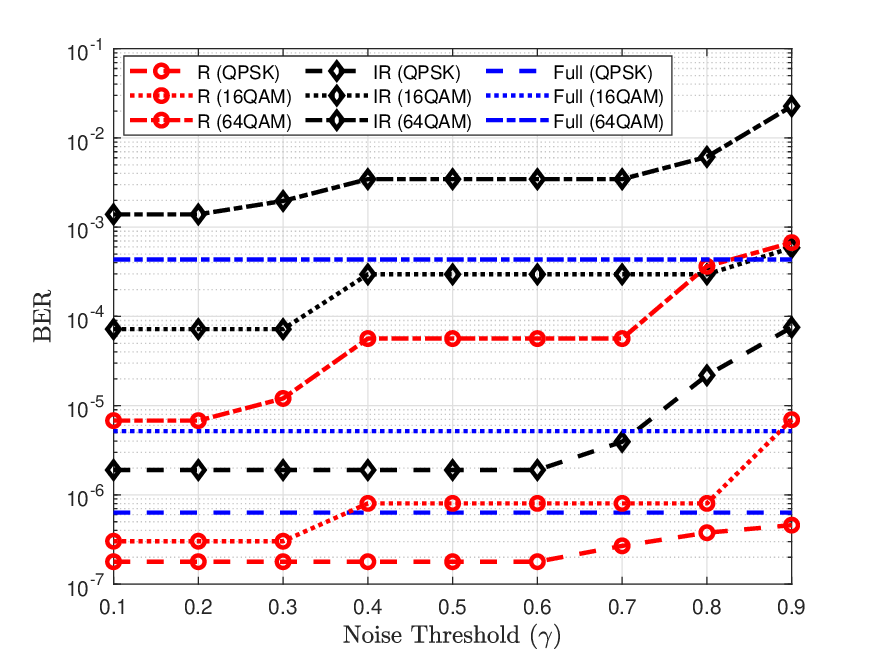}} 
    \caption{Noise distribution and BER threshold analysis for LFS channel model employing QPSK, 16QAM, and 64QAM, respectively.}
    \label{fig:B1}
\end{figure*}

However, we can notice that for higher the modulation order, the number of neglected subcarriers increases. This is because the conventional STA channel estimates at these subcarriers are so noisy, so the STA-FNN model neglects them. It is worth mentioning that, the STA-FNN model treats the conventional STA estimated channels separately and does not consider the time correlation between successive OFDM symbols due to the architectural design of the FNN network. Hence, the channel tracking over time is applied within the conventional STA scheme, where the FNN model is used to capture the frequency correlation of the channel samples as well as coping with the conventional STA estimation error. In this context, the STA-FNN model neglects subcarriers due to two main reasons: (\textit{i}) LFS: The channel variation among the subcarriers is slow, so few subcarriers are required to accurately estimate the channel, as we will discuss in the next Section. (\textit{i}) HFS: Here the channel variation is significant among the subcarriers, thus, the $U$ model should consider more relevant subcarriers to guarantee good channel estimation accuracy. However, this is subject to the condition where the conventional estimated channel at the considered subcarriers is useful and not so noisy. Therefore, in the HFS channel model, more relevant subcarriers are needed and this is shown in generally shifting the noise weight distribution towards zero. However, for higher modulation orders, mainly 64QAM, the neglected subcarriers are huge due to the bad channel estimation quality at these subcarriers. Hence, avoiding them is useful to guarantee BER performance.
We note that the four pilot subcarriers are assigned the lowest noise weight for all the modulation orders. Therefore, the $U$ model is able to classify the pilots as the most relevant subcarriers regardless of the channel's high selectivity and the employed modulation order.

Figure~{\ref{fig:C1}}(b) shows the {\ac{BER}} in terms of $\gamma$ considering SNR = $40$ dB using the HFS channel model. We can notice that ($\gamma = 0.5$, $|\Psi| = 20$), and ($\gamma = 0.5$, $|\Psi| = 11$)  are the best options corresponding to the 16QAM, and 64QAM modulation orders, respectively. Again we can notice that the functionality of the interpretability model in classifying the subcarriers into relevant and irrelevant is based on the context, i.e., the employed modulation order in this case.

\begin{figure*}[t]
    \centering
    \subfigure[Noise weight distribution]{\includegraphics[width=0.32\textwidth]{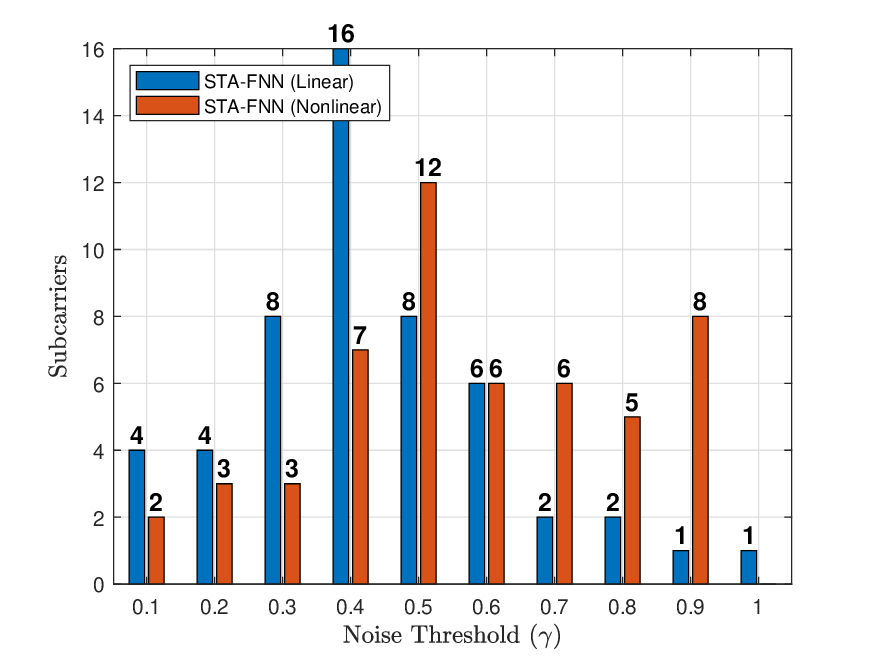}} 
    \subfigure[BER performance vs. noise threshold]{\includegraphics[width=0.32\textwidth]{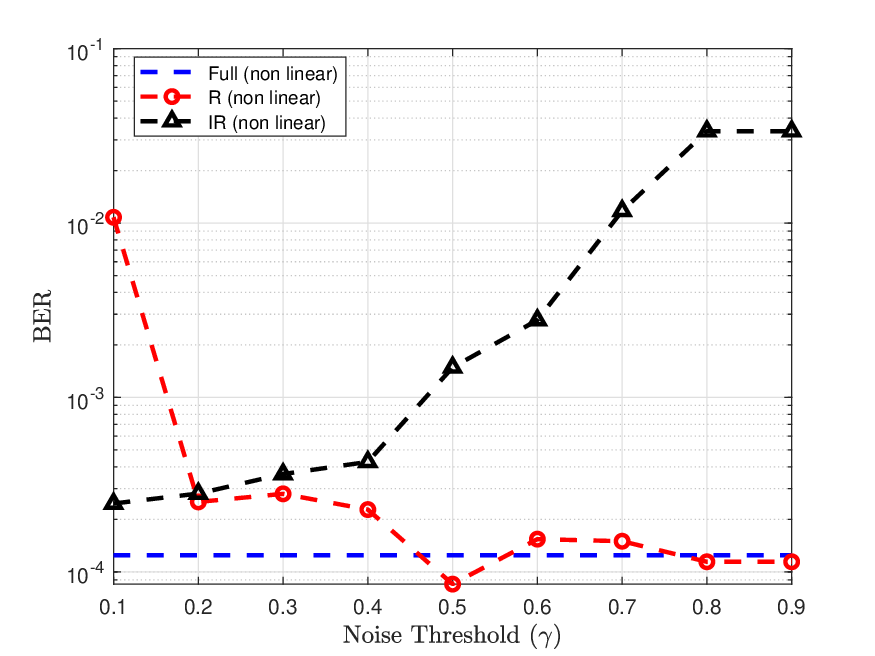}} 
    \subfigure[HFS channel model - IBO = 2dB]{\includegraphics[width=0.32\textwidth]{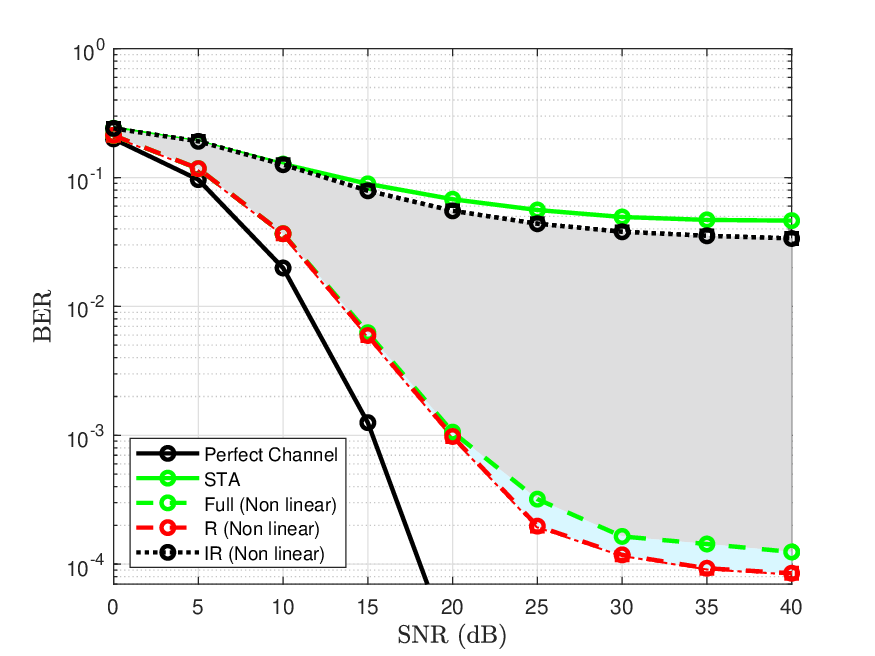}}
    \caption{BER considering HFS channel model employing QPSK and IBO = 2 dB. We note that the highlighted gray and blue areas refer to the BER performance when considering the irrelevant and relevant subcarriers, respectively. }
    \label{fig:HFS_Non}
\end{figure*}

\subsection{\MakeUppercase{Impact of channel frequency selectivity} \label{LFS}}

In this section, we will investigate the performance evaluation using the same methodology of Section~{\ref{mod_order}} but considering the LFS channel model. Figure~\ref{fig:B2} shows the BER results of employing QPSK, 16QAM, and 64QAM modulation orders, respectively. We can notice a significant performance degradation as the modulation order increases which is expected. The nice thing lies in employing the pilot subcarriers only, where the corresponding BER performance improves in comparison to the full case. In other words, the BER performance improvement of employing the pilots in comparison to the full case for 64QAM modulation is higher than that for 16QAM and QPSK modulations, respectively. This is because applying the frequency and time domain averaging in the conventional STA channel estimation is no longer reliable due to high demapping error resulting from the DPA channel estimation~\eqref{eq: DPA_1} that is applied prior to the STA estimation. Similarly to the discussion in Section~\ref{noise_thr}, employing more relevant and irrelevant subcarriers leads to a BER performance degradation in both cases where in the relevant case, the BER performance is approaching the full case, while in the irrelevant case, the performance is going off the full case. 

Figure~\ref{fig:B1}(a) illustrates the noise weight distribution of training the models using different modulation orders. We can notice that distribution is shifted towards one, where the majority of subcarriers are assigned noise weight equal to one. This signifies that these subcarriers are not important for the decision-making methodology of the $U$ model. This is because, in the LFS channel model, the channel presents a smooth variation over the subcarriers, thus, the STA-FNN model needs few subcarriers to accomplish the channel estimation task. Moreover, as the modulation order increases, the noise weight distribution becomes wider, where more subcarriers are assigned more weights. For example, in the 64QM modulation order, it seems that the model needs more subcarriers to preserve good performance, thus the number of subcarriers that are assigned noise weight = 1 decreases. Moreover, in all cases, the model is able to classify pilots as the most relevant subcarriers by assigning them the lowest noise weight regardless of the employed modulation order. The BER vs the noise weight for the considered modulation orders is shown in Figure~{\ref{fig:B1}}(b) where we can notice that considering only the pilots in the LFS channel model is enough, and there is no need to consider any other subcarriers. On the contrary, all the irrelevant subcarrier combinations are worse than the full case in terms of {\ac{BER}} performance. Hence, the absence of the four pilots leads to performance degradation even if the other $|\Psi| = 48$ subcarriers are considered.

\subsection{\MakeUppercase{Impact of RF non-linear distortion}}

In order to further analyze the impact of HPA-induced nonlinearities, we employ QPSK modulation and IBO = 2 dB in the HFS channel models. Figure~{\ref{fig:HFS_Non}} shows the noise weight distribution as well as the BER analysis. It can be noticed that only 2 pilots are assigned the lowest noise weight in comparison to 4 in the linear case. This ensures that the HPA-induced nonlinearities contribute to confusing the subcarrier filtering procedure. However, a slight BER rate performance improvement can be achieved by employing $|\Psi| = 27$ for $\gamma = 0.5$. Therefore, similar insights can be concluded as the linear case where the proposed perturbation based XAI framework is able to filter out the relevant subcarriers while preserving the BER performance when using the full subscribers as an input to the $U$ model.

\begin{figure*}[h]
    \centering
    \subfigure[Noise weight distribution]{\includegraphics[width=0.49\textwidth]{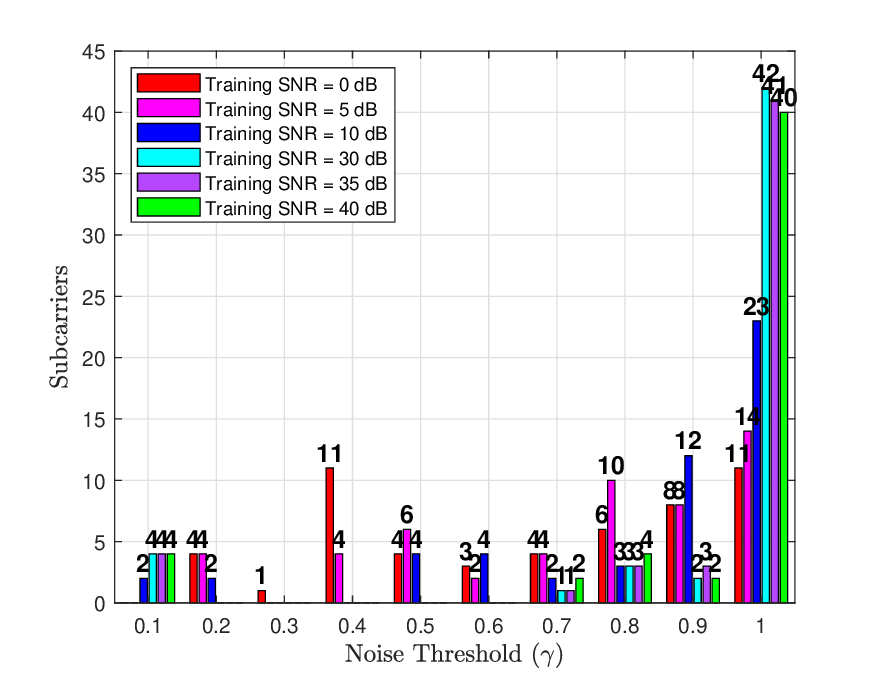}} 
    \subfigure[LFS channel model - QPSK modulation]{\includegraphics[width=0.49\textwidth]{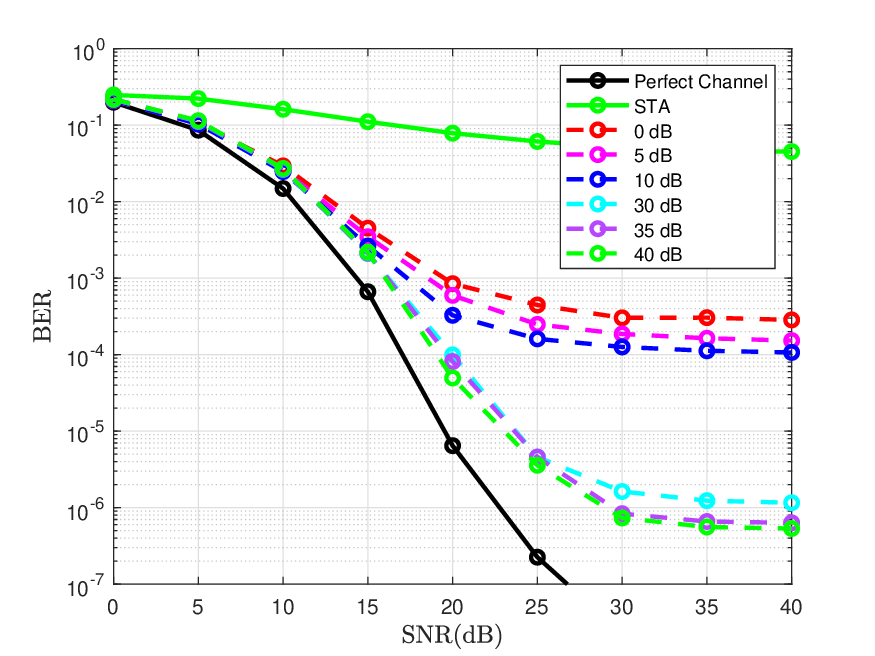}} 
    \caption{Noise distribution and BER performance considering the LFS channel model and QPSK modulation employing different training SNRs.}
    \label{fig:D}
\end{figure*}

\subsection{\MakeUppercase{Impact of training SNR}}

The sensitivity of the $U$ model training, considering different SNR values, is analyzed in this section. \figref{fig:D}(a) shows the noise distribution when considering several training SNRs employing the LFS channel model and QPSK modulation order. Starting by training SNR = $0-5$ dB, we can see that the pilot subcarriers are assigned $0.2$ noise weight and the distribution is flattened along the entire range. This reveals that even though the pilots have accurate channel estimates, due to the dominant impact of AWGN noise, the $U$ model is not able to assign the lowest noise weight to the pilot subcarriers. It is worth mentioning that when training on SNR = $10$ dB, the model starts to identify the pilot subcarriers as the most relevant subcarriers by assigning to two pilots the lowest noise weight, i.e., $0.1$. Moreover, as the training SNR increases, the noise distribution is shifted more towards one, signifying that the model is better identifying the relevant and irrelevant subcarriers. \figref{fig:D}(b) shows the {\ac{BER}} performance when the $U$ model is trained on a specific SNR and tested on the entire SNR range. We can notice that training on higher SNR gives better performance than training on the lower SNR due to the fact the AWGN noise is negligible at high SNRs, thus the $U$ model can learn more efficiently the channel. In addition, the trained model on high SNR can perform well when tested on lower SNRs due to the generalization ability of FNN networks. In conclusion, training on low SNR values leads to a limited performance improvement over the conventional STA channel estimation. Whereas training on high SNR allows the smart feature selection resulting in optimizing the $U$ model input size, as well as significantly improving the BER performance in comparison to the conventional STA channel estimation.

\begin{figure*}[t]
    \centering
    \subfigure[Noise weight distribution]{\includegraphics[width=0.49\textwidth]{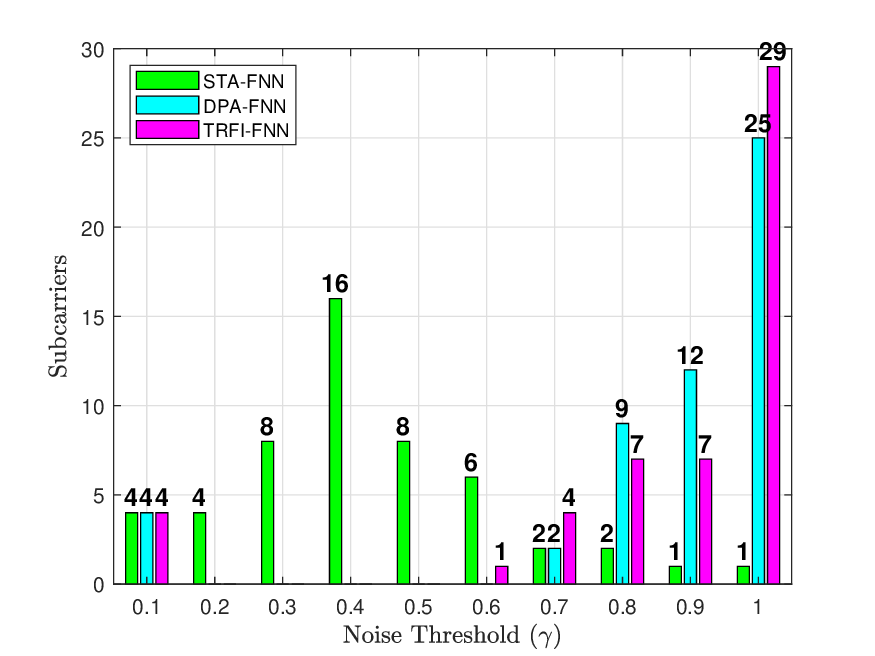}} 
    \subfigure[HFS channel model - QPSK modulation]{\includegraphics[width=0.49\textwidth]{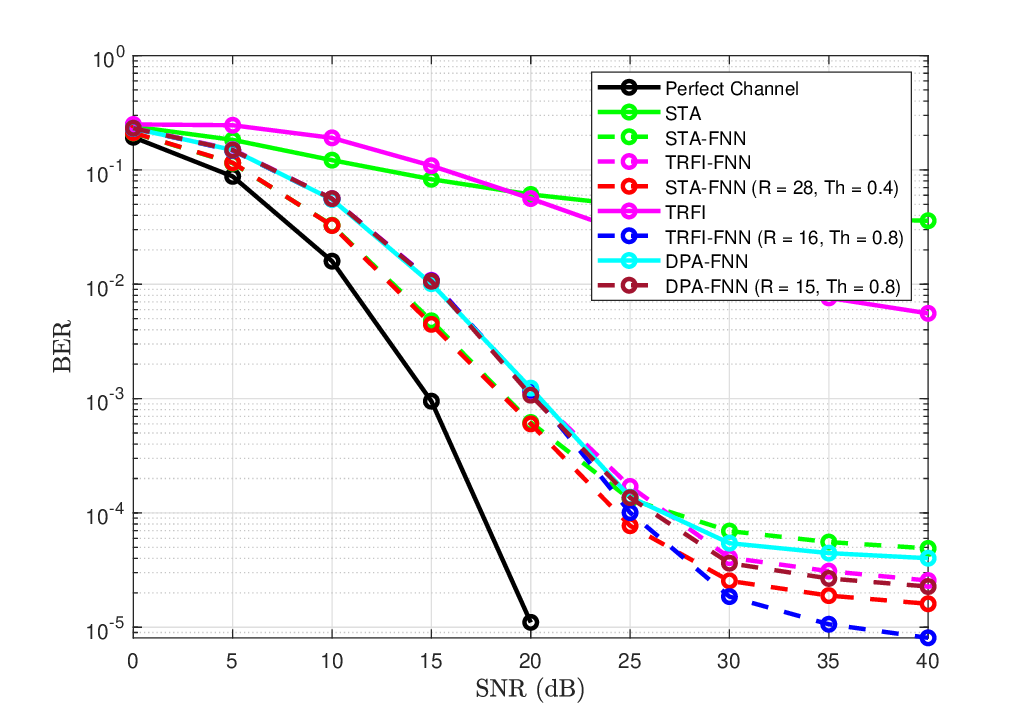}} 
    \caption{Noise distribution and BER performance of the DPA-FNN, STA-FNN, and TRFI-FNN channel estimation schemes.}
    \label{fig:E}
\end{figure*}

\subsection{\MakeUppercase{Impact of conventional channel estimation accuracy}}

To further analyze the impact of the conventional channel estimation, which is implemented prior to the FNN processing on the noise weight distribution, we considered in this section the DPA-FNN~\cite{ref_AE_DNN} and TRFI-FNN~\cite{ref_TRFI_DNN} channel estimation schemes in addition to STA-FNN. We note that we consider the HFS channel model with QPSK modulation in this analysis since it is more challenging.

As we can see from Figure~\ref{fig:D}(b), the conventional TRFI channel estimation outperforms the STA channel estimation in the high SNR region. This is due to the cubic interpolation employed on top of the DPA channel estimation in the TRFI scheme. Similar behavior can be seen with respect to the TRFI-FNN and STA-FNN channel estimators, where the TRFI-FNN also outperforms the STA-FNN in the high SNR region.

Motivated by the fact that the conventional TRFI is better than the conventional STA channel estimation, it is expected that the TRFI-FNN model may neglect more subcarriers than the STA-FNN channel estimation scheme. This is shown in Figure~\ref{fig:D}(a), where we can notice that even though we consider the HFS channel model, the noise distribution of the HFS channel model is still shifted towards one, signifying that the TRFI-FNN requires less relevant subcarriers than the STA-FNN channel estimation scheme in order to preserve the BER performance as the full case, where $|\Psi| = 52$. Similar behavior is recorded for the DPA-FNN channel estimation scheme, where the distribution is also shifted towards one in a close manner as the TRFI-FNN scheme. This is because the conventional TRFI scheme slightly outperforms the DPA channel estimation in the considered scenario. However, the STA-FNN and TRFI-FNN channel estimation schemes outperform the DPA-FNN due to the averaging operations and cubic interpolation employed in the conventional STA and TRFI schemes, respectively.

In this context, we can conclude that as the accuracy of the conventional channel estimation increases, the number of selected important subcarriers decreases, where STA-FNN requires $|\Psi| = 28$ relevant subcarriers, which are greater than the relevant subcarriers required by the TRFI-FNN, i.e.,  $|\Psi| = 16$. This means that the $N$ model is able to induce more noise to the TRFI-FNN input, whereas less noise is induced to the STA-FNN input since it is already noisy.

\begin{table*}[t]
\renewcommand{\arraystretch}{1.6}
\centering
\caption{FLOPS of the optimized FNN architecture employed in the STA-FNN channel estimation scheme.}
\label{table_sbs_flops}
\begin{tabular}{|c|c|c|c|c|c|c|}
\hline
\textbf{FNN Architecture} & Full (15-15-15) & Relevant (15-15-15) & Relevant (15-15) & Relevant (15) & Relevant (10) & Relevant (5) \\ \hline
\textbf{FLOPS}            & 7.52 K          & 4.64 K              & 4.13 K           & 3.62 K        & 2.48 K        & 1.34 K       \\ \hline
\end{tabular}
\end{table*}

\subsection{\MakeUppercase{Computational complexity reduction}}

This section aims to investigate the possibility of optimizing the U model architecture following selecting the most relevant subcarriers so that the BER performance improvement as well as reducing the computational complexity can be achieved. In this context, we considered the same simulation parameters as Section~\ref{noise_thr}, where the pilot subcarriers are fed to the U model with different architectures. The objective is to reduce the computational complexity of the classical STA-FNN model ($15-15-15$) while preserving the BER performance of the best relevant case, i.e., employing only the pilots in the LFS channel model. 

\begin{figure}[t]
    \centering
\includegraphics[width=1\columnwidth]{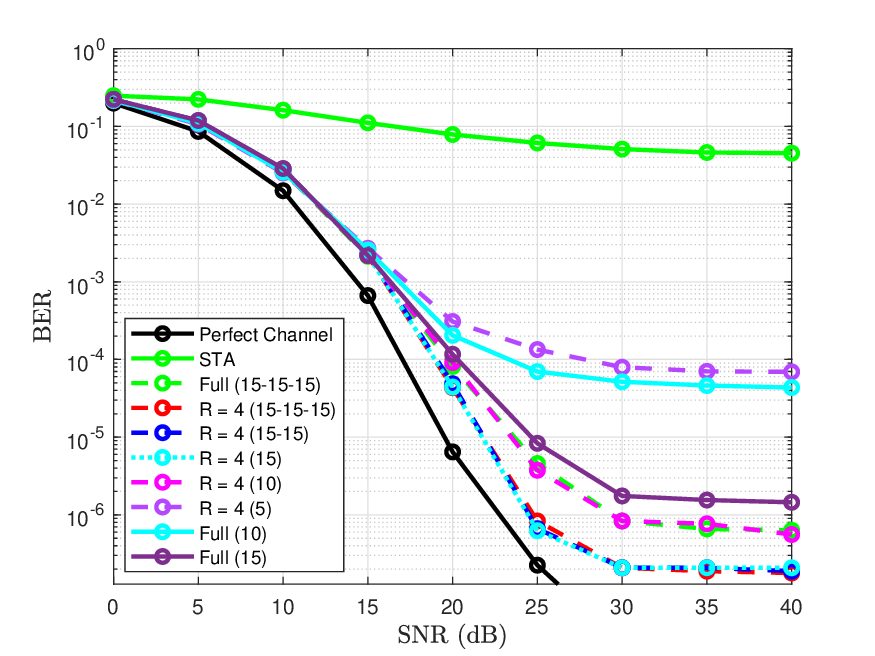}
    \caption{STA-FNN BER performance using different optimized FNN architectures.}
    \label{fig:fnn_optimization}
\end{figure}

Figure~\ref{fig:fnn_optimization} shows the BER performance of different STA-FNN architectures. We can notice that the FNN architecture could be reduced up to one hidden layer with $15$ neurons while preserving the best possible performance. Moreover, decreasing the number of neurons within this architecture to $10$ performs the same as the classical STA-FNN channel estimation scheme, i.e., considering the full subcarriers as inputs with the ($15-15-15$) FNN architecture. However, employing shallow FNN architecture with $5$ neurons is not useful at all, where a significant performance degradation is recorded in comparison to the classical STA-FNN architecture.

The computational complexity of the employed FNNs is computed in terms of the number of FLOPS\footnote{We note that the number of FLOPS are calculated using the pytorch-OpCounter package~\cite{ptflops}.} required by each FNN architecture, as shown in Table~\ref{table_sbs_flops}. Employing the same FNN architecture as the classical STA-FNN one but using the pilot subcarriers as input reduces the complexity by around $1.5\times$ times in comparison to the classical STA-FNN channel estimation scheme. However, further complexity reduction can be recorded by employing a shallow FNN with $15$ neurons, where  $2\times$ times can be achieved in comparison to the classical STA-FNN channel estimation scheme. We would like to mention that similar BER performance can be guaranteed as the classical STA-FNN channel estimator by feeding the four pilots to a shallow FNN architecture with $10$ neurons, where $3\times$ times less computational complexity is required. Finally, we would like to mention that the proposed XAI-CHEST framework resolves the main issues related to the black box DL models by providing interpretability to the model behavior, performance improvement, and computational complexity reduction by selecting the relevant model inputs and optimizing the architecture of the employed FNN model.

\section{\MakeUppercase{Conclusion and Future Perspectives}} \label{conclusions}

Ensuring the transparency and trustworthiness of AI is a crucial need for its efficient and safe deployment in critical applications. In this paper we designed a novel XAI-CHEST framework that provides the interpretability of the FNN models employed in the channel estimation application. The XAI-CHEST framework aims to classify the black-box model inputs into relevant and irrelevant inputs by using a perturbation-based methodology. We developed the theoretical foundations of the XAI-CHEST framework by formalizing the related loss functions. Moreover, the noise threshold fine-tuning optimization problem has been analytically derived. Extensive simulations have been conducted where the results reveal that a trustworthy, optimized, and low-complexity channel estimation scheme can be designed by selecting only the relevant inputs. 

As a future perspective, three main research directions could be established:

\begin{itemize}
    \item XAI-CHEST for RNN-based channel estimation: The functionality of the proposed XAI-CHEST framework is limited since it is not yet adapted to cope with the time variation of the wireless channel. Hence, it is essential to extend the XAI-CHEST framework to deal with RNN-based channel estimation. Thanks to the RNN memory that allows the prediction of the current channel based on previously estimated channels, RNN networks such as \ac{LSTM} and \ac{GRU} are able to perform channel estimation and tracking over time. Therefore, adapting the XAI-CHEST to the RNN-based channel estimation provides better noise allocation that varies over time among the received OFDM symbols. Moreover, investigating the impact of the RNN memory size of the efficiency of the noise allocation could provide further performance-complexity trade-offs.

    \item XAI-CHEST for MIMO-OFDM: The combination of multi-antenna and multi-carrier technologies is a promising technique for ensuring the efficiency of high-speed transmission in wireless communication systems. However, providing performance-complexity trade-offs is crucial for better designing the MIMO-OFDM receiver. In this context, extending the proposed XAI-CHEST for MIMO-OFDM could be beneficial in adapting the size of the considered MIMO system based on the channel correlation and the desired performance.

    \item Gradient-assisted XAI-CHEST framework: The proposed XAI-CHEST framework is based on an external model-agnostic perturbation-based methodology. Hence, the provided model's interpretability is impacted only by studying the influence of the model inputs on its decision, where the internal architecture of the model remains black-box. Therefore, the current XAI-CHEST framework provides a smart input filtering strategy where model-driven optimization is still unexplored. In this context, investigating internal gradient-based XAI schemes~{\cite{10621232}} and integrating them within the XAI-CHEST framework will provide a double optimization strategy that leads to filtering the relevant model inputs, as well as fine-tuning the model architecture where relevant layers and neurons are preserved.
\end{itemize}

% \hl{we will investigate the impact of the channel time-selectivity by adapting the proposed XAI-CHEST framework to the RNN-based channel estimation schemes. 

% Moreover, the extension to multiple-input and multiple-output (MIMO) scenario will be considered. Finally, the gradient-based XAI for wireless communication will be investigated, where we will study the possibility of providing robust interpretation by using the internal architecture of the considered model.}

% \section*
% {ACKNOWLEDGMENT}

% \hl{TBD}

\bibliographystyle{IEEEtran}
\bibliography{ref}

\begin{thebibliography}{10}
\providecommand{\url}[1]{#1}
\csname url@samestyle\endcsname
\providecommand{\newblock}{\relax}
\providecommand{\bibinfo}[2]{#2}
\providecommand{\BIBentrySTDinterwordspacing}{\spaceskip=0pt\relax}
\providecommand{\BIBentryALTinterwordstretchfactor}{4}
\providecommand{\BIBentryALTinterwordspacing}{\spaceskip=\fontdimen2\font plus
\BIBentryALTinterwordstretchfactor\fontdimen3\font minus \fontdimen4\font\relax}
\providecommand{\BIBforeignlanguage}[2]{{%
\expandafter\ifx\csname l@#1\endcsname\relax
\typeout{** WARNING: IEEEtran.bst: No hyphenation pattern has been}%
\typeout{** loaded for the language `#1'. Using the pattern for}%
\typeout{** the default language instead.}%
\else
\language=\csname l@#1\endcsname
\fi
#2}}
\providecommand{\BIBdecl}{\relax}
\BIBdecl

\bibitem{ref_6G}
C.-X. Wang, X.~You, X.~Gao, X.~Zhu, Z.~Li, C.~Zhang, H.~Wang, Y.~Huang, Y.~Chen, H.~Haas, J.~S. Thompson, E.~G. Larsson, M.~D. Renzo, W.~Tong, P.~Zhu, X.~Shen, H.~V. Poor, and L.~Hanzo, ``{On the Road to 6G: Visions, Requirements, Key Technologies, and Testbeds},'' \emph{IEEE Communications Surveys \& Tutorials}, vol.~25, no.~2, pp. 905--974, 2023.

\bibitem{ref_nativeAI}
G.~Liu, Y.~Huang, N.~Li, J.~Dong, J.~Jin, Q.~Wang, and N.~Li, ``{Vision, requirements and network architecture of 6G mobile network beyond 2030},'' \emph{China Communications}, vol.~17, no.~9, pp. 92--104, 2020.

\bibitem{ref_6G_2}
H.~Yang, A.~Alphones, Z.~Xiong, D.~Niyato, J.~Zhao, and K.~Wu, ``{Artificial-Intelligence-Enabled Intelligent 6G Networks},'' \emph{IEEE Network}, vol.~34, no.~6, pp. 272--280, 2020.

\bibitem{ref_latency}
M.~I. {Ashraf}, {Chen-Feng Liu}, M.~{Bennis}, and W.~{Saad}, ``{Towards Low-Latency and Ultra-Reliable Vehicle-to-Vehicle Communication},'' in \emph{2017 European Conference on Networks and Communications (EuCNC)}, 2017, pp. 1--5.

\bibitem{ref_survey}
A.~K. Gizzini and M.~Chafii, ``{A Survey on Deep Learning Based Channel Estimation in Doubly Dispersive Environments},'' \emph{IEEE Access}, vol.~10, pp. 70\,595--70\,619, 2022.

\bibitem{ref_ai_phy_1}
H.~Huang, S.~Guo, G.~Gui, Z.~Yang, J.~Zhang, H.~Sari, and F.~Adachi, ``{Deep Learning for Physical-Layer 5G Wireless Techniques: Opportunities, Challenges and Solutions},'' \emph{IEEE Wireless Communications}, vol.~27, no.~1, pp. 214--222, 2020.

\bibitem{ref_AE_DNN}
S.~{Han}, Y.~{Oh}, and C.~{Song}, ``{A Deep Learning Based Channel Estimation Scheme for IEEE 802.11p Systems},'' in \emph{IEEE International Conference on Communications (ICC)}, 2019, pp. 1--6.

\bibitem{ref_STA_DNN}
A.~K. {Gizzini}, M.~{Chafii}, A.~{Nimr}, and G.~{Fettweis}, ``{Deep Learning Based Channel Estimation Schemes for IEEE 802.11p Standard},'' \emph{IEEE Access}, vol.~8, pp. 113\,751--113\,765, 2020.

\bibitem{ref_TRFI_DNN}
A.~K. Gizzini, M.~Chafii, A.~Nimr, and G.~Fettweis, ``{Joint TRFI and Deep Learning for Vehicular Channel Estimation},'' in \emph{{IEEE GLOBECOM 2020}}, Taipei, Taiwan, Dec. 2020.

\bibitem{ref_wi_cnn}
A.~Karim~Gizzini, M.~Chafii, A.~Nimr, R.~M. Shubair, and G.~Fettweis, ``{CNN Aided Weighted Interpolation for Channel Estimation in Vehicular Communications},'' \emph{IEEE Transactions on Vehicular Technology}, vol.~70, no.~12, pp. 12\,796--12\,811, 2021.

\bibitem{ref_DL_Chest3}
H.~{Ye}, G.~Y. {Li}, and B.~{Juang}, ``{Power of Deep Learning for Channel Estimation and Signal Detection in OFDM Systems},'' \emph{IEEE Wireless Communications Letters}, vol.~7, no.~1, pp. 114--117, 2018.

\bibitem{ref_DL_Chest2}
X.~{Ma}, H.~{Ye}, and Y.~{Li}, ``{Learning Assisted Estimation for Time- Varying Channels},'' in \emph{2018 15th International Symposium on Wireless Communication Systems (ISWCS)}, 2018, pp. 1--5.

\bibitem{ref_DL_Chest1}
Y.~{Yang}, F.~{Gao}, X.~{Ma}, and S.~{Zhang}, ``{Deep Learning-Based Channel Estimation for Doubly Selective Fading Channels},'' \emph{IEEE Access}, vol.~7, pp. 36\,579--36\,589, 2019.

\bibitem{ref_STA}
J.~A. {Fernandez}, K.~{Borries}, L.~{Cheng}, B.~V.~K. {Vijaya Kumar}, D.~D. {Stancil}, and F.~{Bai}, ``{Performance of the 802.11p Physical Layer in Vehicle-to-Vehicle Environments},'' \emph{IEEE Transactions on Vehicular Technology}, vol.~61, no.~1, pp. 3--14, 2012.

\bibitem{ref_TRFI}
{Yoon-Kyeong Kim}, {Jang-Mi Oh}, {Yoo-Ho Shin}, and {Cheol Mun}, ``{Time and Frequency Domain Channel Estimation Scheme for IEEE 802.11p},'' in \emph{17th International IEEE Conference on Intelligent Transportation Systems (ITSC)}, 2014, pp. 1085--1090.

\bibitem{ref_RNN_paper}
A.~K. Gizzini and M.~Chafii, ``{RNN Based Channel Estimation in Doubly Selective Environments},'' \emph{IEEE Transactions on Machine Learning in Communications and Networking}, vol.~2, pp. 1--18, 2024.

\bibitem{ref_DL_CHEST_LSTM}
J.~Pan, H.~Shan, R.~Li, Y.~Wu, W.~Wu, and T.~Q.~S. Quek, ``{Channel Estimation Based on Deep Learning in Vehicle-to-Everything Environments},'' \emph{IEEE Communications Letters}, vol.~25, no.~6, pp. 1891--1895, 2021.

\bibitem{ref_BiRNN}
A.~K. Gizzini and M.~Chafii, ``{Deep Learning Based Channel Estimation in High Mobility Communications Using Bi-RNN Networks},'' in \emph{{ICC 2023 - IEEE International Conference on Communications}}, 2023, pp. 2607--2612.

\bibitem{ref_CNN5}
L.~Li, H.~Chen, H.-H. Chang, and L.~Liu, ``{Deep Residual Learning Meets OFDM Channel Estimation},'' \emph{IEEE Wireless Communications Letters}, vol.~9, no.~5, pp. 615--618, 2020.

\bibitem{ref_CNN6}
D.~Luan and J.~S. Thompson, ``{Channelformer: Attention Based Neural Solution for Wireless Channel Estimation and Effective Online Training},'' \emph{IEEE Transactions on Wireless Communications}, vol.~22, no.~10, pp. 6562--6577, 2023.

\bibitem{ref_trust_XAI}
D.~Kaur, S.~Uslu, A.~Durresi, S.~Badve, and M.~Dundar, ``Trustworthy explainability acceptance: A new metric to measure the trustworthiness of interpretable ai medical diagnostic systems,'' in \emph{Complex, Intelligent and Software Intensive Systems}, L.~Barolli, K.~Yim, and T.~Enokido, Eds.\hskip 1em plus 0.5em minus 0.4em\relax Cham: Springer International Publishing, 2021, pp. 35--46.

\bibitem{nielsen2022robust}
I.~E. Nielsen, D.~Dera, G.~Rasool, R.~P. Ramachandran, and N.~C. Bouaynaya, ``{Robust Explainability: A Tutorial on Gradient-based Attribution Methods for Deep Neural Networks},'' \emph{IEEE Signal Processing Magazine}, vol.~39, no.~4, pp. 73--84, 2022.

\bibitem{arya2019one}
V.~Arya, R.~K. Bellamy, P.-Y. Chen, A.~Dhurandhar, M.~Hind, S.~C. Hoffman, S.~Houde, Q.~V. Liao, R.~Luss, A.~Mojsilovi{\'c} \emph{et~al.}, ``{One explanation does not fit all: A toolkit and taxonomy of ai explainability techniques},'' \emph{arXiv preprint arXiv:1909.03012}, 2019.

\bibitem{ref_xai_6G_1}
W.~Guo, ``{Explainable Artificial Intelligence for 6G: Improving Trust between Human and Machine},'' \emph{IEEE Communications Magazine}, vol.~58, no.~6, pp. 39--45, 2020.

\bibitem{ref_xai_6G_2}
Y.~Wu, G.~Lin, and J.~Ge, ``{Knowledge-Powered Explainable Artificial Intelligence for Network Automation toward 6G},'' \emph{IEEE Network}, vol.~36, no.~3, pp. 16--23, 2022.

\bibitem{ref_xai_6G_ran_1}
B.~Brik, H.~Chergui, L.~Zanzi, F.~Devoti, A.~Ksentini, M.~S. Siddiqui, X.~Costa-Pérez, and C.~Verikoukis, ``{A Survey on Explainable AI for 6G O-RAN: Architecture, Use Cases, Challenges and Research Directions},'' 2023.

\bibitem{ref_xai_6G_ran_2}
F.~Rezazadeh, H.~Chergui, L.~Alonso, and C.~Verikoukis, ``{SliceOps: Explainable MLOps for Streamlined Automation-Native 6G Networks},'' 2023.

\bibitem{ref_xai_6G_ran_3}
F.~Rezazadeh, H.~Chergui, and J.~Mangues-Bafalluy, ``{Explanation-Guided Deep Reinforcement Learning for Trustworthy 6G RAN Slicing},'' in \emph{2023 IEEE International Conference on Communications Workshops (ICC Workshops)}, 2023, pp. 1026--1031.

\bibitem{lundberg2017unified}
S.~M. Lundberg and S.-I. Lee, ``{A unified approach to interpreting model predictions},'' \emph{Advances in neural information processing systems}, vol.~30, 2017.

\bibitem{ref_xai_res_alloc_1}
P.~Barnard, I.~Macaluso, N.~Marchetti, and L.~A. DaSilva, ``{Resource Reservation in Sliced Networks: An Explainable Artificial Intelligence (XAI) Approach},'' in \emph{{ICC 2022 - IEEE International Conference on Communications}}, 2022, pp. 1530--1535.

\bibitem{ref_xai_res_alloc_2}
A.-D. Marcu, S.~K. Gowtam~Peesapati, J.~Moysen~Cortes, S.~Imtiaz, and J.~Gross, ``{Explainable Artificial Intelligence for Energy-Efficient Radio Resource Management},'' in \emph{2023 IEEE Wireless Communications and Networking Conference (WCNC)}, 2023, pp. 1--6.

\bibitem{ref_xai_res_alloc_3}
N.~Khan, S.~Coleri, A.~Abdallah, A.~Celik, and A.~M. Eltawil, ``{Explainable and Robust Artificial Intelligence for Trustworthy Resource Management in 6G Networks},'' \emph{IEEE Communications Magazine}, pp. 1--7, 2023.

\bibitem{ref_xai_iot_1}
S.~K. Jagatheesaperumal, Q.-V. Pham, R.~Ruby, Z.~Yang, C.~Xu, and Z.~Zhang, ``{Explainable AI Over the Internet of Things (IoT): Overview, State-of-the-Art and Future Directions},'' \emph{IEEE Open Journal of the Communications Society}, vol.~3, pp. 2106--2136, 2022.

\bibitem{ref_xai_3}
M.~Zolanvari, Z.~Yang, K.~Khan, R.~Jain, and N.~Meskin, ``{TRUST XAI: Model-Agnostic Explanations for AI With a Case Study on IIoT Security},'' \emph{IEEE Internet of Things Journal}, vol.~10, no.~4, pp. 2967--2978, 2023.

\bibitem{ribeiro2016should}
M.~T. Ribeiro, S.~Singh, and C.~Guestrin, ``{Why Should I Trust You? Explaining the Predictions of Any Classifier},'' in \emph{Proceedings of the 22nd ACM SIGKDD international conference on knowledge discovery and data mining}, 2016, pp. 1135--1144.

\bibitem{10368353}
A.~K. Gizzini, Y.~Medjahdi, A.~J. Ghandour, and L.~Clavier, ``{Towards Explainable AI for Channel Estimation in Wireless Communications},'' \emph{IEEE Transactions on Vehicular Technology}, pp. 1--6, 2023.

\bibitem{ref_IEEE_Spec}
A.~Abdelgader and L.~Wu, ``{The Physical Layer of the IEEE 802.11 p WAVE Communication Standard: The Specifications and Challenges},'' in \emph{{The Physical Layer of the IEEE 802.11 p WAVE Communication Standard: The Specifications and Challenges}}, vol.~2, 10 2014.

\bibitem{bussgang1952crosscorrelation}
J.~J. Bussgang, ``Crosscorrelation functions of amplitude-distorted gaussian signals,'' 1952.

\bibitem{9743925}
A.~K. Gizzini and M.~Chafii, ``Low complex methods for robust channel estimation in doubly dispersive environments,'' \emph{IEEE Access}, vol.~10, pp. 34\,321--34\,339, 2022.

\bibitem{boyd2004convex}
S.~P. Boyd and L.~Vandenberghef, \emph{Convex optimization}.\hskip 1em plus 0.5em minus 0.4em\relax Cambridge university press, 2004.

\bibitem{r19}
I.~{Sen} and D.~W. {Matolak}, ``{Vehicle–Vehicle Channel Models for the 5-GHz Band},'' \emph{IEEE Transactions on Intelligent Transportation Systems}, vol.~9, no.~2, pp. 235--245, 2008.

\bibitem{ref_LS_DNN}
A.~K. {Gizzini}, M.~{Chafii}, A.~{Nimr}, and G.~{Fettweis}, ``{Enhancing Least Square Channel Estimation Using Deep Learning},'' in \emph{2020 IEEE 91st Vehicular Technology Conference (VTC2020-Spring)}, 2020, pp. 1--5.

\bibitem{ptflops}
\BIBentryALTinterwordspacing
V.~Sovrasov. (2018-2023) {ptflops: a Flops Counting Tool for Neural Networks in Pytorch Framework}. [Online]. Available: \url{https://github.com/sovrasov/flops-counter.pytorch}
\BIBentrySTDinterwordspacing

\bibitem{10621232}
A.~K. Gizzini, Y.~Medjahdi, and M.~B. Mabrouk, ``{GRACE: Gradient-based XAI Scheme for Channel Estimation in Wireless Communications},'' in \emph{2024 IEEE International Mediterranean Conference on Communications and Networking (MeditCom)}, 2024, pp. 572--577.

\end{thebibliography}

\vfill\pagebreak

\end{document}